\documentclass[journal,compsoc]{IEEEtran}
\usepackage{amsmath}
\usepackage{amssymb}
\usepackage{array}
\usepackage{bm}
\usepackage{multirow}
\usepackage{graphicx} % more modern
\usepackage{subfigure}
\usepackage{url}
\usepackage{cite}
\usepackage{algorithm}
\usepackage{algorithmic}
\usepackage{caption}
\ifCLASSOPTIONcompsoc
\else
\fi
\ifCLASSINFOpdf
\else
\fi
\ifCLASSOPTIONcompsoc
  \usepackage[caption=false,font=normalsize,labelfont=sf,textfont=sf]{subfig}
\else
  \usepackage[caption=false,font=footnotesize]{subfig}
\fi

\ifCLASSOPTIONcaptionsoff
  \usepackage[nomarkers]{endfloat}
 \let\MYoriglatexcaption\caption
 \renewcommand{\caption}[2][\relax]{\MYoriglatexcaption[#2]{#2}}
\fi
\begin{document}
%
% paper title
% can use linebreaks \\ within to get better formatting as desired
% Do not put math or special symbols in the title.
\title{Iterated Support Vector Machines for Distance Metric Learning}
%
%
% author names and IEEE memberships
% note positions of commas and nonbreaking spaces ( ~ ) LaTeX will not break
% a structure at a ~ so this keeps an author's name from being broken across
% two lines.
% use \thanks{} to gain access to the first footnote area
% a separate \thanks must be used for each paragraph as LaTeX2e's \thanks
% was not built to handle multiple paragraphs
%
%
%\IEEEcompsocitemizethanks is a special \thanks that produces the bulleted
% lists the Computer Society journals use for "first footnote" author
% affiliations. Use \IEEEcompsocthanksitem which works much like \item
% for each affiliation group. When not in compsoc mode,
% \IEEEcompsocitemizethanks becomes like \thanks and
% \IEEEcompsocthanksitem becomes a line break with idention. This
% facilitates dual compilation, although admittedly the differences in the
% desired content of \author between the different types of papers makes a
% one-size-fits-all approach a daunting prospect. For instance, compsoc
% journal papers have the author affiliations above the "Manuscript
% received ..."  text while in non-compsoc journals this is reversed. Sigh.

\author{Wangmeng Zuo,~\IEEEmembership{Member,~IEEE,}
        Faqiang Wang,
        David Zhang,~\IEEEmembership{Fellow,~IEEE,}
        Liang Lin,~\IEEEmembership{Member,~IEEE,}
        Yuchi Huang,~\IEEEmembership{Member,~IEEE,}
        Deyu Meng, and
        Lei Zhang,~\IEEEmembership{Senior Member,~IEEE}        % <-this % stops a space
\IEEEcompsocitemizethanks{\IEEEcompsocthanksitem W. Zuo and F. Wang are with the School of Computer Science and Technology, Harbin Institute of Technology, Harbin, 150001, China. (e-mail: cswmzuo@gmail.com; tshfqw@163.com)
\IEEEcompsocthanksitem D. Zhang and L. Zhang are with the Department of Computing, the Hong Kong Polytechnic University, Kowloon, Hong Kong. (e-mail: csdzhang@comp.polyu.edu.hk; cslzhang@comp.polyu.edu.hk)% <-this % stops a space
\IEEEcompsocthanksitem L. Lin is with the School of Super-computing, Sun Yat-Sen University, Guangzhou, 510275, China. (e-mail: linliang@ieee.org)% <-this % stops a space
\IEEEcompsocthanksitem Y. Huang is with the NEC Laboratories China, Beijing, 100084, China. (e-mail: huang\_yuchi@nec.cn)% <-this % stops a space
\IEEEcompsocthanksitem D. Meng is with the Institute of Information and System Sciences, Faculty of Mathematics and Statistics, Xi'an Jiaotong University, Xi'an, 710049, China. (e-mail: dymeng@mail.xjtu.edu.cn)}% <-this % stops a space
\thanks{Manuscript received XXX; revised XXX.}}
\IEEEtitleabstractindextext{%
\begin{abstract}
Distance metric learning aims to learn from the given training data a valid distance metric, with which the similarity between data samples can be more effectively evaluated for classification. Metric learning is often formulated as a convex or nonconvex optimization problem, while many existing metric learning algorithms become inefficient for large scale problems. In this paper, we formulate metric learning as a kernel classification problem, and solve it by iterated training of support vector machines (SVM). The new formulation is easy to implement, efficient in training, and tractable for large-scale problems. Two novel metric learning models, namely Positive-semidefinite Constrained Metric Learning (PCML) and Nonnegative-coefficient Constrained Metric Learning (NCML), are developed. Both PCML and NCML can guarantee the global optimality of their solutions. Experimental results on UCI dataset classification, handwritten digit recognition, face verification and person re-identification demonstrate that the proposed metric learning methods achieve higher classification accuracy than state-of-the-art methods and they are significantly more efficient in training.
\end{abstract}

% Note that keywords are not normally used for peerreview papers.
}

% make the title area
\maketitle

% To allow for easy dual compilation without having to reenter the
% abstract/keywords data, the \IEEEtitleabstractindextext text will
% not be used in maketitle, but will appear (i.e., to be "transported")
% here as \IEEEdisplaynontitleabstractindextext when the compsoc
% or transmag modes are not selected <OR> if conference mode is selected
% - because all conference papers position the abstract like regular
% papers do.
\IEEEdisplaynontitleabstractindextext
% \IEEEdisplaynontitleabstractindextext has no effect when using
% compsoc or transmag under a non-conference mode.

% For peer review papers, you can put extra information on the cover
% page as needed:
 %\ifCLASSOPTIONpeerreview
 %\begin{center} \bfseries EDICS Category: 3-BBND \end{center}
 %\fi
%
% For peerreview papers, this IEEEtran command inserts a page break and
% creates the second title. It will be ignored for other modes.
\IEEEpeerreviewmaketitle

\section{Introduction}
% The very first letter is a 2 line initial drop letter followed
% by the rest of the first word in caps.
%
% form to use if the first word consists of a single letter:
% \IEEEPARstart{A}{demo} file is ....
%
% form to use if you need the single drop letter followed by
% normal text (unknown if ever used by IEEE):
% \IEEEPARstart{A}{}demo file is ....
%
% Some journals put the first two words in caps:
% \IEEEPARstart{T}{his demo} file is ....
%
% Here we have the typical use of a "T" for an initial drop letter
% and "HIS" in caps to complete the first word.
\IEEEPARstart{D}{istance} metric learning aims to train a valid distance metric which can enlarge the distances between samples of different classes and reduce the distances between samples of the same class \cite{bellet2013}. Metric learning is closely related to $k$-Nearest Neighbor ($k$-NN) classification \cite{weinberger2009LMNNJMLR}, clustering \cite{xing2002distance}, ranking \cite{mcfee2010metric,lim2013robust}, feature extraction \cite{zhang2011learning} and support vector machine (SVM) \cite{xu2013distance}, and has been widely applied to face recognition \cite{guillaumin2009LDML}, person re-identification \cite{kostinger2012large,li2013learning}, image retrieval \cite{hoi2008semi,yang2010boosting}, activity recognition \cite{tran2008human}, document classification \cite{lebanon2006metric}, and link prediction \cite{shaw2011learning}, etc. One popular metric learning approach is the Mahalanobis distance metric learning, which is to learn a linear transformation matrix $\mathbf{L}$ or a matrix $\mathbf{M}=\mathbf{L}^T\mathbf{L}$ from the training data. Given two samples $\mathbf{x}_i$ and $\mathbf{x}_j$, the Mahalanobis distance between them is defined as:
\begin{equation}
\begin{aligned}
d_{\mathbf{M}}^{2}\left( {{\mathbf{x}}_{i}},{{\mathbf{x}}_{j}} \right)& =\left\| \mathbf{L}({{\mathbf{x}}_{i}}-{{\mathbf{x}}_{j}}) \right\|_{2}^{2}\\
& ={{\left( {{\mathbf{x}}_{i}}-{{\mathbf{x}}_{j}} \right)}^{T}}\mathbf{M}\left( {{\mathbf{x}}_{i}}-{{\mathbf{x}}_{j}} \right).
\end{aligned}
\end{equation}
\par
To satisfy the nonnegative property of a distance metric, $\mathbf{M}$ should be positive semidefinite (PSD). According to which one of $\mathbf{M}$ and $\mathbf{L}$ is learned, Mahalanobis distance metric learning methods can be grouped into two categories. Methods that learn $\mathbf{L}$, including neighborhood components analysis (NCA) \cite{goldberger2004NCA}, large margin components analysis (LMCA) \cite{torresani2006large} and neighborhood repulsed metric learning (NRML) \cite{lu2014neighborhood}, are mostly formulated as nonconvex optimization problems, which are solved by gradient descent based optimizers. Taking the PSD constraint into account, methods that learn $\mathbf{M}$, including large margin nearest neighbor (LMNN) \cite{weinberger2005LMNNNIPS} and maximally collapsing metric learning (MCML) \cite{globerson2005MCML}, are mostly formulated as convex semidefinite programming (SDP) problems, which can be optimized by standard SDP solvers \cite{weinberger2005LMNNNIPS}, projected gradient \cite{xing2002distance}, Boosting-like \cite{shen2012positive}, or Frank-Wolfe \cite{ying2012distance} algorithms. Davis \textit{et al.} \cite{davis2007information} proposed an information-theoretic metric learning (ITML) model with an iterative Bregman projection algorithm, which does not need projections onto the PSD cone. Besides, the use of online solvers for metric learning has been discussed in \cite{mensink2012metric,kostinger2012large,checkik2010large}.
\par
On the other hand, kernel methods \cite{belkin2006manifold,andrews2002support,evgeniou2004regularized,pekalska2009kernel,anand2014semisupervised,maji2013efficient} have been widely studied in many learning tasks, e.g., semi-supervised learning, multiple instance learning, multitask learning, etc. Kernel learning methods, such as support vector machine (SVM), exhibit good generalization performance. There are many open resources on kernel classification methods, and a variety of toolboxes and libraries have been released \cite{vapnik1995nature,chang2011,Platt1999,tsang2005core,bordes2007,teo2007scalable,shalev2011pegasos}. It is thus important to investigate the connections between metric learning and kernel classification and explore how to utilize the kernel classification resources in the research and development of metric learning methods.
\begin{table}[t]
\caption{Summary of main abbreviations}
%\vspace{-5mm}
\label{Abbrev}
\renewcommand{\arraystretch}{1.3}
%\vskip 0in
\begin{center}
\begin{tabular}{c|p{6cm}}
\hline
\bfseries Abbreviation & \bfseries Full Name \\
\hline
PSD & Positive semidefinite (matrix)\\
SDP & Semidefinite programming\\
$k$-NN & $k$-nearest neighbor (classification)\\
KKT & Karush-Kuhn-Tucker (condition)\\
SVM & Support vector machine\\
LMCA\cite{torresani2006large} & Large margin components analysis\\
LMNN\cite{weinberger2009LMNNJMLR} & Large margin nearest neighbor\\
NCA\cite{goldberger2004NCA} & Neighborhood components analysis\\
MCML\cite{globerson2005MCML} & Maximally collapsing metric learning\\
ITML\cite{davis2007information} & Information-theoretic metric learning\\
LDML\cite{guillaumin2009LDML} & Logistic discriminant metric learning\\
DML-eig\cite{ying2012distance} & Distance metric learning with eigenvalue optimization\\
PLML\cite{wang2012PLML} & Parametric local metric learning\\
KISSME\cite{kostinger2012large} & Keep it simple and straightforward metric learning\\
PCML & Positive-semidefinite constrained metric learning\\
NCML & Nonnegative-coefficient constrained metric learning\\
\hline
\end{tabular}
\end{center}
%\vskip -0in
\vspace{-5mm}
\end{table}
\par
In this paper, we propose a novel formulation of metric learning by casting it as a kernel classification problem, which allows us to effectively and efficiently learn distance metrics by iterated training of SVM. The off-the-shelf SVM solvers such as LibSVM \cite{chang2011} can be employed to solve the metric learning problem. Specifically, we propose two novel methods to bridge metric learning with the well-developed SVM techniques, and they are easy to implement. First, we propose a Positive-semidefinite Constrained Metric Learning (PCML) model, which can be solved via iterating between PSD projection and dual SVM learning. Second, by re-parameterizing the matrix $\mathbf{M}$, we transform the PSD constraint into a nonnegative coefficient constraint and consequently propose a Nonnegative-coefficient Constrained Metric Learning (NCML) model, which can be solved by iterated learning of two SVMs. Both PCML and NCML have globally optimal solutions, and our extensive experiments on UCI dataset classification, handwritten digit recognition, face verification and person re-identification clearly demonstrate the effectiveness of them.
\par
The remainder of this paper is organized as follows. Section \ref{relatedwork} reviews the related works. Section \ref{pcml} presents the PCML model and the optimization algorithm. Section \ref{ncml} presents the model and algorithm of NCML. Section \ref{experimentalresults} presents the experimental results, and Section \ref{conclusion} concludes the paper.
\par
The main abbreviations used in this paper are summarized in Table 1.

\section{Related Work}
\label{relatedwork}
\par
Compared with nonconvex metric learning models \cite{goldberger2004NCA,torresani2006large,niu2012information}, convex formulation of metric learning \cite{weinberger2009LMNNJMLR,xing2002distance,globerson2005MCML,shen2012positive,ying2012distance} has drawn increasing attentions due to its desired properties such as global optimality. Most convex metric learning models can be formulated as SDP or quadratic SDP problems. Standard SDP solvers, however, are inefficient for metric learning, especially when the size of training samples is big or the feature dimension is high. Therefore, customized optimization algorithm needs to be developed for each specific metric learning model. For LMNN, Weinberger \textit{et al.} developed an efficient solver based on the sub-gradient descent and the active set techniques \cite{weinberger2008fast}. In ITML, Davis \textit{et al.} \cite{davis2007information} suggested an iterative Bregman projection algorithm. Iterative projected gradient descent method \cite{xing2002distance,jin2009regularized} has been widely employed for metric learning but it requires an eigenvalue decomposition in each iteration. Other algorithms such as block-coordinate descent \cite{qi2009efficient}, smooth optimization \cite{ying2009sparse}, and Frank-Wolfe \cite{ying2012distance} have also been studied for metric learning. Unlike the customized algorithms, in this work we formulate metric learning as a kernel classification problem and solve it using the off-the-shelf SVM solvers, which can guarantee the global optimality and the PSD property of the learned $\mathbf{M}$, and is easy to implement and efficient in training.
\par
Another line of work aims to develop metric learning algorithms by solving the Lagrange dual problems. Shen \textit{et al.} derived the Lagrange dual of the exponential loss based metric learning model, and proposed a boosting-like approach, namely BoostMetric, where the matrix $\mathbf{M}$ is learned as a linear positive combination of rank-one matrices \cite{shen2012positive,shen2009positive}. MetricBoost \cite{bi2011} and FrobMetric \cite{shen2011scalable,shen2014efficient} were further proposed to improve the performance of BoostMetric. Liu and Vemuri incorporated two regularization terms in the duality for robust metric learning \cite{liu2012robust}. Note that BoostMetric \cite{shen2012positive,shen2009positive}, MetricBoost \cite{bi2011}, and FrobMetric \cite{shen2011scalable} are proposed for metric learning with triplet constraints, whereas in many applications such as verification, only pairwise constraints are available in the training stage.
\par
Several SVM-based metric learning approaches \cite{nguyen2008metric,brunner2012,do2012metric,wang2014kernel} have also been proposed. Using SVM, Nguyen and Guo \cite{nguyen2008metric} formulated metric learning as a quadratic semidefinite programming problem, and suggested a projected gradient descent algorithm. The formulations of the proposed PCML and NCML in this work are different from the model in \cite{nguyen2008metric}, and they are solved by the dual problems with the off-the-shelf SVM solvers. Brunner \MakeLowercase{\textit{et al.}} \cite{brunner2012} proposed a pairwise SVM method to learn a dissimilarity function rather than a distance metric. Different from \cite{brunner2012}, the proposed PCML and NCML learn a distance metric and the matrix $\mathbf{M}$ is constrained to be a PSD matrix. Do \MakeLowercase{\textit{et al.}} \cite{do2012metric} studied SVM from a metric learning perspective and presented an improved variant of SVM classification. Wang \MakeLowercase{\textit{et al.}} \cite{wang2014kernel} developed a kernel classification framework for metric learning and proposed two learning models which can be efficiently implemented by the standard SVM solvers. However, they adopted a two-step greedy strategy to solve the models and neglected the PSD constraint in the first step. In this work, the proposed PCML and NCML models have different formulations from \cite{wang2014kernel}, and their solutions are globally optimal.

\begin{flushleft}
\section{Positive-semidefinite Constrained Metric Learning (PCML)}
\label{pcml}
\end{flushleft}
Denote by $\left\{ \left. \left( {{\mathbf{x}}_{i}},{{y}_{i}} \right) \right|i=1,2,\cdots ,N \right\}$ a training set, where ${{\mathbf{x}}_{i}}\in {{\mathbb{R}}^{d}}$ is the $i$th training sample, and $y_i$ is the class label of $\mathbf{x}_i$. The Mahalanobis distance between $\mathbf{x}_i$ and $\mathbf{x}_j$ can be equivalently written as:
\begin{equation}
\begin{aligned}
d_{\mathbf{M}}^{2}\left( {{\mathbf{x}}_{i}},{{\mathbf{x}}_{j}} \right)& =\operatorname{tr}\left( {{\mathbf{M}}^{T}}({{\mathbf{x}}_{i}}-{{\mathbf{x}}_{j}}){{({{\mathbf{x}}_{i}}-{{\mathbf{x}}_{j}})}^{T}} \right)\\
& =\left\langle \mathbf{M},\left( {{\mathbf{x}}_{i}}-{{\mathbf{x}}_{j}} \right){{\left( {{\mathbf{x}}_{i}}-{{\mathbf{x}}_{j}} \right)}^{T}} \right\rangle,
\end{aligned}
\end{equation}	
where $\mathbf{M}$ is a PSD matrix, $\left\langle \mathbf{A},\mathbf{B} \right\rangle =\operatorname{tr}\left( {{\mathbf{A}}^{T}}\mathbf{B} \right)$ is defined as the Frobenius inner product of two matrices $\mathbf{A}$ and $\mathbf{B}$, and $\operatorname{tr}(\bullet )$ stands for the matrix trace operator. For each pair of $\mathbf{x}_i$ and $\mathbf{x}_j$, we define a matrix ${{\mathbf{X}}_{ij}}=({{\mathbf{x}}_{i}}-{{\mathbf{x}}_{j}}){{({{\mathbf{x}}_{i}}-{{\mathbf{x}}_{j}})}^{T}}$. With ${{\mathbf{X}}_{ij}}$, the Mahalanobis distance can be rewritten as $d_{\mathbf{M}}^{2}\left( {{\mathbf{x}}_{i}},{{\mathbf{x}}_{j}} \right)=\left\langle \mathbf{M},{{\mathbf{X}}_{ij}} \right\rangle $.
\subsection{PCML and Its Dual Problem}
Let $\mathcal{S}=\{\left(\mathbf{x}_i,\mathbf{x}_j\right): \text{the } \text{class } \text{labels } \text{of } \mathbf{x}_i \text{ and } \mathbf{x}_j \text{ are the }$ $\text{same}\}$  be the set of similar pairs, and let $\mathcal{D}=\{\left(\mathbf{x}_i,\mathbf{x}_j\right): \text{the class labels of } \mathbf{x}_i \text{ and } \mathbf{x}_j \text{ are different}\}$ be the set of dissimilar pairs. By introducing an indicator variable $h_{ij}$
\begin{equation}
\begin{aligned}
{{h}_{ij}}=\left\{ \begin{matrix}
   1,\text{    if (}{{\mathbf{x}}_{i}}\text{, }{{\mathbf{x}}_{j}}\text{)}\in \mathcal{D}  \\
   -1,\text{ if (}{{\mathbf{x}}_{i}}\text{, }{{\mathbf{x}}_{j}}\text{)}\in \mathcal{S},  \\
\end{matrix} \right.
\end{aligned}
\end{equation}
the PCML model can be formulated as:
\begin{equation}
\begin{aligned}
  \underset{\mathbf{M},b,\boldsymbol{\xi }}{\mathop{\min }}\, \quad & \frac{1}{2}\left\| \mathbf{M} \right\|_{F}^{2}+C\sum\nolimits_{i,j}{{{\xi }_{ij}}} \\
 \text{s}\text{.t}\text{.} \quad & {{h}_{ij}}\left( \left\langle \mathbf{M},{{\mathbf{X}}_{ij}} \right\rangle +b \right)\ge 1-{{\xi }_{ij}}, {{\xi }_{ij}}\ge 0,\ \forall i,j \\
 & \mathbf{M}\succcurlyeq 0,
\end{aligned}
\end{equation}
where ${{\xi }_{ij}}$ denotes the slack variables, $b$ denotes the bias, and ${{\left\| \centerdot  \right\|}_{F}}$ denotes the Frobenius norm.
\par
The PCML model defined above is convex and can be solved using the standard SDP solvers. However, the high complexity of general-purpose interior-point SDP solver makes it only suitable for small-scale problems. In order to improve the efficiency, in the following we first analyze the Lagrange duality of the PCML model, and then propose an algorithm to iterate between SVM training and PSD projection to learn the Mahalanobis distance metric.
\par
By introducing the Lagrange multipliers $\boldsymbol{\lambda }$ and a PSD matrix $\mathbf{Y}$, the Lagrange dual of the problem in (4) can be formulated as:
\begin{equation}
\begin{aligned}
  \underset{\boldsymbol{\lambda },\mathbf{Y}}{\mathop{\max }}\, \quad & -\frac{1}{2}\left\| \sum\nolimits_{i,j}{{{\lambda }_{ij}}{{h}_{ij}}{{\mathbf{X}}_{ij}}}+\mathbf{Y} \right\|_{F}^{2}+\sum\nolimits_{i,j}{{{\lambda }_{ij}}} \\
 \text{s}\text{.t}\text{.} \quad & \sum\nolimits_{i,j}{{{\lambda }_{ij}}{{h}_{ij}}}=0, 0\le {{\lambda }_{ij}}\le C,\ \forall i,j,\quad \mathbf{Y}\succcurlyeq 0.
\end{aligned}
\end{equation}
Please refer to \textbf{Appendix A} for the detailed derivation of the dual problem. Based on the Karush-Kuhn-Tucker (KKT) conditions, the matrix $\mathbf{M}$ can be obtained by
\begin{equation}
\begin{aligned}
\mathbf{M}=\sum\nolimits_{i,j}{{{\lambda }_{ij}}{{h}_{ij}}{{\mathbf{X}}_{ij}}}+\mathbf{Y}.
\end{aligned}
\end{equation}
The strong duality allows us to first solve the equivalent dual problem in (5) and then obtain the matrix $\mathbf{M}$ by (6). However, due to the PSD constraint $\mathbf{Y}\succcurlyeq 0$, the problem in (5) is still difficult to optimize.
\subsection{Alternative Optimization Algorithm}
\par
To solve the dual problem efficiently, we propose an optimization approach by updating $\boldsymbol{\lambda }$ and $\mathbf{Y}$ alternatively. Given $\mathbf{Y}$, we introduce a new variable $\boldsymbol{\eta}$ with ${{\eta }_{ij}}=1-{{h}_{ij}}\left\langle {{\mathbf{X}}_{ij}},\mathbf{Y} \right\rangle =1-{{h}_{ij}}{{\left( {{\mathbf{x}}_{i}}-{{\mathbf{x}}_{j}} \right)}^{T}}\mathbf{Y}\left( {{\mathbf{x}}_{i}}-{{\mathbf{x}}_{j}} \right)$, and the subproblem on $\boldsymbol{\lambda }$ can be formulated as:
\begin{equation}
\begin{aligned}
\label{PCMLlambda}
  & \underset{\boldsymbol{\lambda }}{\mathop{\max }}\, -\frac{1}{2}\sum\nolimits_{i,j}{\sum\nolimits_{k,l}{{{\lambda }_{ij}}{{\lambda }_{kl}}{{h}_{ij}}{{h}_{kl}}\left\langle {{\mathbf{X}}_{ij}},{{\mathbf{X}}_{kl}} \right\rangle }}+\sum\nolimits_{i,j}{{{\eta }_{ij}}{{\lambda }_{ij}}} \\
 & \quad\text{s.t.}\quad \sum\nolimits_{i,j}{{{\lambda }_{ij}}{{h}_{ij}}}=0, 0\le {{\lambda }_{ij}}\le C,\quad \forall i,j.
\end{aligned}
\end{equation}
\par
The subproblem (\ref{PCMLlambda}) is a QP problem. We can define a kernel function of sample pairs as follows:
\begin{equation}
\begin{aligned}
  K\left( \left( {{\mathbf{x}}_{i}},{{\mathbf{x}}_{j}} \right),\left( {{\mathbf{x}}_{k}},{{\mathbf{x}}_{l}} \right) \right) & =\left\langle {{\mathbf{X}}_{ij}},{{\mathbf{X}}_{kl}} \right\rangle \\
  & ={{\left( {{\left( {{\mathbf{x}}_{i}}-{{\mathbf{x}}_{j}} \right)}^{T}}\left( {{\mathbf{x}}_{k}}-{{\mathbf{x}}_{l}} \right) \right)}^{2}}.
\end{aligned}
\end{equation}
Substituting (8) into (7), the subproblem on $\boldsymbol{\lambda}$ becomes a kernel-based classification problem, and can be efficiently solved by using the existing SVM solvers such as LibSVM \cite{chang2011}.
Given $\boldsymbol{\lambda }$, the subproblem on $\mathbf{Y}$ can be formulated as the projection of a matrix onto the convex cone of PSD matrices:
\begin{equation}
\begin{aligned}
  & \underset{\mathbf{Y}}{\mathop{\min }}\,\quad \left\| \mathbf{Y}-{{\mathbf{Y}}_{0}} \right\|_{F}^{2},\quad  \text{s}\text{.t}\text{.}\quad \mathbf{Y}\succcurlyeq 0,
\end{aligned}
\end{equation}
where ${{\mathbf{Y}}_{0}}=-\sum\nolimits_{i,j}{{{\lambda }_{ij}}{{h}_{ij}}{{\mathbf{X}}_{ij}}}$. Through the eigen-decomposition of $\mathbf{Y}_0$, i.e., ${{\mathbf{Y}}_{0}}=\mathbf{U\Lambda }{{\mathbf{U}}^{T}}$ and $\boldsymbol{\Lambda }$ is the diagonal matrix of eigenvalues, the solution to the subproblem on $\mathbf{Y}$ can be explicitly expressed as $\mathbf{Y}=\mathbf{U}\boldsymbol{\Lambda}_{+}{{\mathbf{U}}^{T}}$, where $\boldsymbol{\Lambda}_{+}=\max \left( \boldsymbol{\Lambda },\mathbf{0} \right)$. Finally, the PCML algorithm is summarized in \textbf{Algorithm \ref{alg:PCML}}.
\begin{algorithm}[tb]
   \caption{Algorithm of PCML}
   \label{alg:PCML}
\begin{algorithmic}
   \STATE {\bfseries Input:} $\mathcal{S}=\{\left(\mathbf{x}_i,\mathbf{x}_j\right): \text{the class labels of } \mathbf{x}_i \text{ and } \mathbf{x}_j$ $\text{ are the same}\},$  $\mathcal{D}=\{\left(\mathbf{x}_i,\mathbf{x}_j\right): \text{the class labels of } \mathbf{x}_i$ $\text{ and } \mathbf{x}_j \text{ are different}\},$ $\text{and}$ $h_{ij}.$
   \STATE {\bfseries Output:} $\mathbf{M}$.
   \STATE {\bfseries Initialize} ${{\mathbf{Y}}^{\left( 0 \right)}}$, $t\leftarrow 0$.
   \REPEAT
   \STATE 1. Update ${{\mathbf{\eta }}^{\left( t+1 \right)}}$ with $\eta _{ij}^{(t+1)}=1-{{h}_{ij}}\left\langle {{\mathbf{X}}_{ij}},{{\mathbf{Y}}^{(t)}} \right\rangle $.
   \STATE 2. Update ${{\boldsymbol{\lambda }}^{\left( t+1 \right)}}$ by solving the subproblem (7) using an SVM solver.
   \STATE 3. Update $\mathbf{Y}_{0}^{(t+1)}=-\sum\nolimits_{i,j}{\lambda _{ij}^{(t+1)}{{h}_{ij}}{{\mathbf{X}}_{ij}}}$.
   \STATE 4. Update ${{\mathbf{Y}}^{(t+1)}}=\mathbf{U}^{(t+1)}\boldsymbol{\Lambda}_{+}^{(t+1)}{{\mathbf{U}^{(t+1)}}^{T}}$, where $\mathbf{Y}_{0}^{(t+1)}=\mathbf{U}^{(t+1)}\mathbf{\Lambda }^{(t+1)}{{\mathbf{U}^{(t+1)}}^{T}}$ and $\boldsymbol{\Lambda}_{+}^{(t+1)}=\max \left( \boldsymbol{\Lambda }^{(t+1)},\mathbf{0} \right)$.
   \STATE 5. $t\leftarrow t+1$.
   \UNTIL{convergence}
   \STATE $\mathbf{M}=\sum\nolimits_{i,j}{{{\lambda }_{ij}^{(t-1)}}{{h}_{ij}}{{\mathbf{X}}_{ij}}}+\mathbf{Y}^{(t-1)}$.
   \STATE {\bfseries return} $\mathbf{M}$
\end{algorithmic}
\end{algorithm}
\subsection{Optimality Condition}
As shown in \cite{csisz1984information, gunawardana2005convergence}, the general alternating minimization approach will converge. By alternatively updating $\boldsymbol{\lambda }$ and $\mathbf{Y}$, the proposed algorithm can reach the global optimum of the problems in (4) and (5).
\par
The optimality condition of the proposed algorithm can be checked by the duality gap in each iteration, which is defined as the difference between the primal and dual objective values:
\begin{equation}
\begin{aligned}
\text{DualGap}_\text{PCML}^{(n)}= & \frac{1}{2}\left\| {{\mathbf{M}}^{(n)}} \right\|_{F}^{2}+C\sum\nolimits_{i,j}{\xi _{ij}^{(n)}}-\sum\nolimits_{i,j}{\lambda _{ij}^{(n)}}\\
& +\frac{1}{2}\left\| \sum\nolimits_{i,j}{\lambda _{ij}^{(n)}{{h}_{ij}}{{\mathbf{X}}_{ij}}}+{{\mathbf{Y}}^{(n)}} \right\|_{F}^{2},
\end{aligned}
\end{equation}
where ${{\mathbf{M}}^{(n)}}$, $\boldsymbol{\xi }_{{}}^{(n)}$, $\boldsymbol{\lambda }_{{}}^{(n)}$, and ${{\mathbf{Y}}^{(n)}}$ are feasible primal and dual variables, and $\text{DualGap}_{\text{PCML}}^{\left( n \right)}$ is the duality gap in the $n$th iteration. According to (6), we can derive that
\begin{equation}
\begin{aligned}
{{\mathbf{M}}^{\left( n \right)}}=\sum\nolimits_{i,j}{\lambda _{ij}^{\left( n \right)}{{h}_{ij}}{{\mathbf{X}}_{ij}}}+{{\mathbf{Y}}^{\left( n \right)}}={{\mathbf{Y}}^{\left( n \right)}}-\mathbf{Y}_{0}^{\left( n \right)}.
\end{aligned}
\end{equation}
As shown in Subsection 3.2, $\mathbf{Y}_{0}^{\left( n \right)}={{\mathbf{U}}^{\left( n \right)}}{{\mathbf{\Lambda }}^{\left( n \right)}}{{\mathbf{U}}^{\left( n \right)}}^{T}$, ${{\mathbf{Y}}^{\left( n \right)}}={{\mathbf{U}}^{\left( n \right)}}\mathbf{\Lambda }_{+}^{\left( n \right)}{{\mathbf{U}}^{\left( n \right)}}^{T}$, and hence ${{\mathbf{M}}^{\left( n \right)}}={{\mathbf{U}}^{\left( n \right)}}\mathbf{\Lambda }_{-}^{\left( n \right)}{{\mathbf{U}}^{\left( n \right)}}^{T}$, where $\mathbf{\Lambda }_{-}^{\left( n \right)}=\mathbf{\Lambda }_{+}^{\left( n \right)}-{{\mathbf{\Lambda }}^{\left( n \right)}}$. Thus, $\left\| {{\mathbf{M}}^{\left( n \right)}} \right\|_{F}^{2}$ can be computed by
\begin{equation}
\begin{aligned}
\left\| {{\mathbf{M}}^{\left( n \right)}} \right\|_{F}^{2} & =\operatorname{tr}\left( {{\mathbf{M}}^{\left( n \right)}}^{T}{{\mathbf{M}}^{\left( n \right)}} \right)\\
& =\operatorname{tr}\left( {{\mathbf{U}}^{\left( n \right)}}\mathbf{\Lambda }_{-}^{\left( n \right)}{{\mathbf{U}}^{\left( n \right)}}^{T}{{\mathbf{U}}^{\left( n \right)}}\mathbf{\Lambda }_{-}^{\left( n \right)}{{\mathbf{U}}^{\left( n \right)}}^{T} \right)\\
& =\operatorname{tr}\left( {{\mathbf{U}}^{\left( n \right)}}\mathbf{\Lambda }_{-}^{\left( n \right)2}{{\mathbf{U}}^{\left( n \right)}}^{T} \right) =\operatorname{tr}\left( \mathbf{\Lambda }_{-}^{\left( n \right)2} \right).
\end{aligned}
\end{equation}
Substituting (11) and (12) into (10), the duality gap of PCML can be obtained as follows
\begin{equation}
\begin{aligned}
\text{DualGap}_{\text{PCML}}^{\left( n \right)}=C\sum\nolimits_{i,j}{\xi _{ij}^{\left( n \right)}}-\sum\nolimits_{i,j}{\lambda _{ij}^{\left( n \right)}}+\operatorname{tr}\left( \mathbf{\Lambda }{{_{-}^{\left( n \right)}}^{2}} \right).
\end{aligned}
\end{equation}
\par
\begin{figure}[!t]
\centering
\includegraphics[width=0.6\columnwidth]{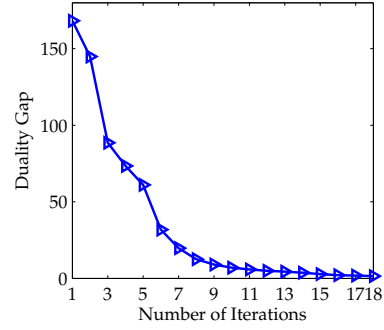}
\caption{Duality gap vs. number of iterations on the \emph{PenDigits} dataset for PCML.}
\label{Dualgap_PCML}
\end{figure}

Based on the KKT conditions of the PCML dual problem in (5), $\xi _{ij}^{\left( n \right)}$ can be obtained by
\begin{equation}
\xi _{ij}^{\left( n \right)}=\left\{ \begin{aligned}
  & 0, \forall \lambda _{ij}^{\left( n \right)}<C \\
 & {{\left[ 1-{{h}_{ij}}\left( \left\langle {{\mathbf{M}}^{\left( n \right)}},{{\mathbf{X}}_{ij}} \right\rangle +{{b}^{\left( n \right)}} \right) \right]}_{+}}, \forall \lambda _{ij}^{\left( n \right)}=C,
\end{aligned} \right.
\end{equation}
where
\begin{equation}
\begin{aligned}
{{b}^{\left( n \right)}}=\frac{1}{{{h}_{ij}}}-\left\langle {{\mathbf{M}}^{\left( n \right)}},{{\mathbf{X}}_{ij}} \right\rangle, \forall 0<\lambda _{ij}^{\left( n \right)}<C.
\end{aligned}
\end{equation}
Please refer to \textbf{Appendix A} for the detailed derivation of $\xi _{ij}^{\left( n \right)}$ and ${{b}^{\left( n \right)}}$. The duality gap is always nonnegative and approaches to zero when the primal problem is convex. Thus, it can be used as the termination condition of the algorithm. Fig. \ref{Dualgap_PCML} plots the curve of duality gap versus the number of iterations on the \emph{PenDigits} dataset by PCML. One can see that the duality gap converges to zero in less than 20 iterations and our algorithm will reach the global optimum. In \textbf{Algorithm \ref{alg:PCML}}, we adopt the following termination condition:
\begin{equation}
\begin{aligned}
\text{DualGap}_{\text{PCML}}^{\left( t \right)}<\varepsilon \cdot \text{DualGap}_{\text{PCML}}^{\left( 1 \right)},
\end{aligned}
\end{equation}
where $\varepsilon $ is a small constant and we set $\varepsilon =0.01$ in the experiment.\\
\subsection{Remarks}
{\bf Warm-start:} In the updating of $\boldsymbol{\lambda }$, we adopt a simple warm-start strategy. We use the solution of the previous iteration as the initialization of the next iteration. Since the previous solution can serve as a good guess, warm-start results in significant improvement in efficiency.
\par
{\bf Construction of pairwise constraints:} Based on the training set, we can introduce $N^2$ pairwise constraints in total. However, in practice we only need to choose a subset of pairwise constraints to reduce the computational cost. For each sample, we find its $k$ nearest neighbors to construct similar pairs and its $k$ farthest neighbors to construct dissimilar pairs. Thus, we only need $2kN$ pairwise constraints. By this strategy, we can reduce the scale of pairwise constraints from $O\left(N^2\right)$ to $O\left(kN\right)$. Since $k$ is usually small constant (=1$\sim$3) in practice, the computational cost of metric learning is much reduced. Similar strategy for constructing pairwise or triplet constraints can be found in \cite{weinberger2009LMNNJMLR, hoi2008semi}.
\par
{\bf Computational Complexity:} We use the LibSVM library for SVM training. The computational complexity of SMO-type algorithms \cite{Platt1999} is $O(k^2N^2d)$. For PSD projection, the complexity of conventional SVD algorithms is $O(d^3)$.
\begin{flushleft}
\section{Nonnegative-coefficient Constrained Metric Learning (NCML)}
\label{ncml}
\end{flushleft}
Given a set of rank-1 PSD matrices ${{\mathbf{M}}_{t}}={{\mathbf{m}}_{t}}\mathbf{m}_{t}^{T} \left( t=\text{ 1},\cdots ,T \right)$, a linear combination of ${{\mathbf{M}}_{t}}$ is defined as $\mathbf{M}=\sum\nolimits_{t}{{{\alpha }_{t}}{{\mathbf{M}}_{t}}}$, where ${{\alpha }_{t}}$ is the scalar combination coefficient. One can easily prove the following \textbf{Theorem 1}.
\par
\newtheorem{theorem}{Theorem}
\begin{theorem}
If the scalar coefficient ${{\alpha }_{t}}\ge 0,\ \forall t$, the matrix $\mathbf{M}=\sum\nolimits_{t}{\alpha_t\mathbf{M}_t}$ is a PSD matrix, where ${{\mathbf{M}}_{t}}={{\mathbf{m}}_{t}}\mathbf{m}_{t}^{T}$ is a rank-1 PSD matrix.
\end{theorem}

\begin{IEEEproof}
Denote by $\mathbf{u}\in {{\mathbb{R}}^{d}}$ a random vector. Based on the expression of $\mathbf{M}$, we have:
\[
\begin{aligned}
  {{\mathbf{u}}^{T}}\mathbf{Mu} & ={{\mathbf{u}}^{T}}\left( \sum\nolimits_{t}{{{\alpha }_{t}}{{\mathbf{m}}_{t}}\mathbf{m}_{t}^{T}} \right)\mathbf{u} \\
 & =\sum\nolimits_{t}{{{\alpha }_{t}}{{\mathbf{u}}^{T}}{{\mathbf{m}}_{t}}\mathbf{m}_{t}^{T}\mathbf{u}}  =\sum\nolimits_{t}{{{\alpha }_{t}}{{\left( {{\mathbf{u}}^{T}}{{\mathbf{m}}_{t}} \right)}^{2}}}.
\end{aligned}
\]
Since ${{\left( {{\mathbf{u}}^{T}}{{\mathbf{m}}_{t}} \right)}^{2}}\ge 0$ and ${{\alpha }_{t}}\ge 0,\ \forall t$, we have ${{\mathbf{u}}^{T}}\mathbf{Mu}\ge 0$. Therefore, $\mathbf{M}$ is a PSD matrix.
\end{IEEEproof}
\subsection{NCML and Its Dual Problem}
Motivated by \textbf{Theorem 1}, we propose to transform the PSD constraint in (4) by re-parameterizing the distance metric $\mathbf{M}$, and develop a nonnegative-coefficient constrained metric learning (NCML) method to learn the PSD matrix $\mathbf{M}$. Given the training data $\mathcal{S}$ and $\mathcal{D}$, a rank-1 PSD matrix $\mathbf{X}_{ij}$ can be constructed for each pair $\left(\mathbf{x}_i,\mathbf{x}_j\right)$. By assuming that the learned matrix should be the linear combination of $\mathbf{X}_{ij}$ with the nonnegative coefficient constraint, the NCML model can be formulated as:
\begin{equation}
\begin{aligned}
  & \underset{\mathbf{M},b,\boldsymbol{\alpha},\boldsymbol{\xi }}{\mathop{\min }}\,\quad \frac{1}{2}\left\| \mathbf{M} \right\|_{F}^{2}+C\sum\nolimits_{i,j}{{{\xi }_{ij}}} \\
 & \text{s}\text{.t}\text{.}\quad {{h}_{ij}}\left( \left\langle \mathbf{M},{{\mathbf{X}}_{ij}} \right\rangle +b \right)\ge 1-{{\xi }_{ij}},\quad {{\alpha }_{ij}}\ge 0,\ {{\xi }_{ij}}\ge 0,\ \forall i,j \\
  & \quad \quad \mathbf{M}=\sum\nolimits_{i,j}{{{\alpha }_{ij}}{{\mathbf{X}}_{ij}}}.
\end{aligned}
\end{equation}
By substituting $\mathbf{M}$ with $\sum\nolimits_{i,j}{{{\alpha }_{ij}}}{{\mathbf{X}}_{ij}}$, we reformulate the NCML model as follows:
\begin{equation}
\begin{aligned}
  & \underset{\boldsymbol{\alpha },b,\boldsymbol{\xi }}{\mathop{\min }}\,\quad \frac{1}{2}\sum\nolimits_{i,j}{\sum\nolimits_{k,l}{{{\alpha }_{ij}}{{\alpha }_{kl}}\left\langle {{\mathbf{X}}_{ij}},{{\mathbf{X}}_{kl}} \right\rangle }}+C\sum\nolimits_{i,j}{{{\xi }_{ij}}} \\
 & \text{s}\text{.t}\text{.}\quad {{h}_{ij}}\left( \sum\nolimits_{k,l}{{{\alpha }_{kl}}\left\langle {{\mathbf{X}}_{ij}},{{\mathbf{X}}_{kl}} \right\rangle }+b \right)\ge 1-{{\xi }_{ij}} \\
 & \quad \quad {{\alpha }_{ij}}\ge 0,\ {{\xi }_{ij}}\ge 0,\ \forall i,j.
\end{aligned}
\end{equation}
\par
By introducing the Lagrange multipliers $\boldsymbol{\eta }$ and $\boldsymbol{\beta }$, the Lagrange dual of the primal problem in (18) can be formulated as:
\begin{equation}
\begin{aligned}
  \underset{\boldsymbol{\eta },\boldsymbol{\beta }}{\mathop{\max }}\,\ & -\frac{1}{2}\sum\nolimits_{i,j}{\sum\nolimits_{k,l}{\left( {{\beta }_{ij}}{{h}_{ij}}+{{\eta }_{ij}} \right)\left( {{\beta }_{kl}}{{h}_{kl}}+{{\eta }_{kl}} \right)\left\langle {{\mathbf{X}}_{ij}},{{\mathbf{X}}_{kl}} \right\rangle }}\\
  & +\sum\nolimits_{i,j}{{{\beta }_{ij}}} \\
 \text{s}\text{.t}\text{.}\quad & \sum\nolimits_{k,l}{{{\eta }_{kl}}\left\langle {{\mathbf{X}}_{ij}},{{\mathbf{X}}_{kl}} \right\rangle }\ge 0,\ 0\le {{\beta }_{ij}}\le C,\ \forall i,j \\
 & \sum\nolimits_{i,j}{{{\beta }_{ij}}{{h}_{ij}}}=0.
\end{aligned}
\end{equation}
\par
Please refer to \textbf{Appendix B} for the detailed derivation of the dual problem. Based on the KKT conditions, the coefficient ${{\alpha }_{ij}}$ can be obtained by:
\begin{equation}
\begin{aligned}
{{\alpha }_{ij}}={{\beta }_{ij}}{{h}_{ij}}+{{\eta }_{ij}}.
\end{aligned}
\end{equation}
Thus, we can first solve the above dual problem, and then obtain the matrix $\mathbf{M}$ by
\begin{equation}
\begin{aligned}
\mathbf{M}=\sum\nolimits_{i,j}{({{\beta }_{ij}}{{h}_{ij}}+{{\eta }_{ij}}}){{\mathbf{X}}_{ij}}.
\end{aligned}
\end{equation}
\subsection{Optimization Algorithm}
There are two groups of variables, $\boldsymbol{\eta }$ and $\boldsymbol{\beta }$, in problem (19). We adopt an alternative optimization approach to solve them. First, given $\boldsymbol{\eta }$, the variables ${\beta }_{ij}$ can be solved as follows:
\begin{equation}
\begin{aligned}
  \underset{\boldsymbol{\beta }}{\mathop{\max }}\,\ & -\frac{1}{2}\sum\nolimits_{i,j}{\sum\nolimits_{k,l}{{{\beta }_{ij}}{{\beta }_{kl}}{{h}_{ij}}{{h}_{kl}}\left\langle {{\mathbf{X}}_{ij}},{{\mathbf{X}}_{kl}} \right\rangle }}+\sum\nolimits_{i,j}{{{\delta }_{ij}}{{\beta }_{ij}}} \\
 \text{s}\text{.t}\text{.} \quad & 0\le {{\beta }_{ij}}\le C,\ \forall i,j, \sum\nolimits_{i,j}{{{\beta }_{ij}}{{h}_{ij}}}=0,
\end{aligned}
\end{equation}
where $\boldsymbol{\delta }$ is the variable with ${{\delta }_{ij}}=\left( 1-{{h}_{ij}}\sum\nolimits_{kl}{{{\eta }_{kl}}\left\langle {{\mathbf{X}}_{ij}},{{\mathbf{X}}_{kl}} \right\rangle } \right)$. Clearly, the subproblem on $\boldsymbol{\beta }$ is exactly the dual problem of SVM, and it can be efficiently solved by any standard SVM solvers, e.g., LibSVM \cite{chang2011}.
\par
Given $\boldsymbol{\beta }$, the subproblem on $\boldsymbol{\eta }$ can be formulated as follows:
\begin{equation}
\begin{aligned}
  & \underset{\boldsymbol{\eta }}{\mathop{\min }}\,\quad \frac{1}{2}\sum\nolimits_{i,j}{\sum\nolimits_{k,l}{{{\eta }_{ij}}{{\eta }_{kl}}\left\langle {{\mathbf{X}}_{ij}},{{\mathbf{X}}_{kl}} \right\rangle }}+\sum\nolimits_{i,j}{{{\eta }_{ij}}{{\gamma }_{ij}}} \\
 & \text{s}\text{.t}\text{.}\quad \sum\nolimits_{k,l}{{{\eta }_{ij}}\left\langle {{\mathbf{X}}_{ij}},{{\mathbf{X}}_{kl}} \right\rangle }\ge 0,\ \forall i,j,
\end{aligned}
\end{equation}
where ${{\gamma }_{ij}}=\sum\nolimits_{kl}{{{\beta }_{kl}}{{h}_{kl}}\left\langle {{\mathbf{X}}_{ij}},{{\mathbf{X}}_{kl}} \right\rangle }$. To simplify the subproblem on $\boldsymbol{\eta }$, we derive the Lagrange dual of (23) based on the KKT condition:
\begin{equation}
\begin{aligned}
{{\eta }_{ij}}={{\mu }_{ij}}-{{h}_{ij}}{{\beta }_{ij}},\quad \forall i,j,
\end{aligned}
\end{equation}
where $\boldsymbol{\mu }$ is the Lagrange dual multiplier. The Lagrange dual problem of (23) is formulated as follows:
\begin{equation}
\begin{aligned}
  & \underset{\boldsymbol{\mu }}{\mathop{\max }}\,\quad -\frac{1}{2}\sum\nolimits_{i,j}{\sum\nolimits_{k,l}{{{\mu }_{ij}}{{\mu }_{kl}}\left\langle {{\mathbf{X}}_{ij}},{{\mathbf{X}}_{kl}} \right\rangle }}+\sum\nolimits_{i,j}{{{\gamma }_{ij}}}{{\mu }_{ij}} \\
 & \text{s}\text{.t}\text{.}\quad {{\mu }_{ij}}\ge 0,\forall i,j.
\end{aligned}
\end{equation}
Please refer to \textbf{Appendix C} for the detailed derivation. Clearly, problem (25) is a simpler QP problem than (23), which can be efficiently solved by the standard SVM solvers.
\par
By alternatively updating $\boldsymbol{\mu }$ and $\boldsymbol{\beta }$, we can solve the NCML dual problem (19). After obtaining the optimal solutions of $\boldsymbol{\mu }$ and $\boldsymbol{\beta }$, the optimal solution of $\boldsymbol{\alpha }$ in problem (18) can be obtained by
\begin{equation}
\begin{aligned}
{{\alpha }_{ij}}={{\mu }_{ij}},\quad \forall i,j.
\end{aligned}
\end{equation}
We then have $\mathbf{M}=\sum\nolimits_{ij}{{{\alpha }_{ij}}}{{\mathbf{X}}_{ij}}$. The NCML algorithm is summarized in \textbf{Algorithm \ref{alg:NCML}}.
\begin{algorithm}[tb]
   \caption{Algorithm of NCML}
   \label{alg:NCML}
\begin{algorithmic}
   \STATE {\bfseries Input:} Training set $\left\{ \left( {{\mathbf{x}}_{i}},{{\mathbf{x}}_{j}} \right),{{h}_{ij}} \right\}$.
   \STATE {\bfseries Output:} The matrix $\mathbf{M}$.
   \STATE {\bfseries Initialize} ${{\boldsymbol{\eta }}^{\left( 0 \right)}}$ with small random values, $t\leftarrow 0$.
   \REPEAT
   \STATE 1. Update ${{\boldsymbol{\delta }}^{\left( t+1 \right)}}$ with $\boldsymbol{\delta }_{ij}^{\left( t+1 \right)}=\left( 1-{{h}_{ij}}\sum\nolimits_{kl}{\eta _{kl}^{\left( t \right)}\left\langle {{\mathbf{X}}_{ij}},{{\mathbf{X}}_{kl}} \right\rangle } \right)$.
   \STATE 2. Update ${{\boldsymbol{\beta }}^{\left( t+1 \right)}}$ by solving the subproblem (15) using an SVM solver.
   \STATE 3. Update ${{\boldsymbol{\gamma }}^{\left( t+1 \right)}}$ with $\gamma _{ij}^{\left( t+1 \right)}=\sum\nolimits_{kl}{\beta _{kl}^{\left( t+1 \right)}{{h}_{kl}}\left\langle {{\mathbf{X}}_{ij}},{{\mathbf{X}}_{kl}} \right\rangle }$.
   \STATE 4. Update ${{\boldsymbol{\mu }}^{\left( t+1 \right)}}$ by solving the subproblem (18) using an SVM solver.
   \STATE 5. Update ${{\boldsymbol{\eta }}^{\left( t+1 \right)}}$ with $\eta _{ij}^{\left( t+1 \right)}\leftarrow \mu _{ij}^{\left( t+1 \right)}-{{h}_{ij}}\beta _{ij}^{\left( t+1 \right)}$.
   \STATE 6. $t\leftarrow t+1$.
   \UNTIL{convergence}
   \STATE $\mathbf{M}=\sum\nolimits_{ij}{\mu _{ij}^{\left( t \right)}}{{\mathbf{X}}_{ij}}$.
   \STATE {\bfseries return} $\mathbf{M}$
\end{algorithmic}
\end{algorithm}
\par
Analogous to PCML, the updating of $\boldsymbol{\beta }$ and $\boldsymbol{\mu }$ in NCML can be speeded up by using the warm-start strategy. As shown in Fig. \ref{Dualgap_NCML}, the proposed NCML algorithm will converge in 10$\thicksim$15 iterations.
\subsection{Optimality Condition}
We check the duality gap of NCML to investigate the optimality condition of it. From the primal and dual objectives in (18) and (19), the NCML duality gap in the $n$th iteration is
\begin{equation}
\begin{aligned}
  & \text{DualGap}_{\text{NCML}}^{\left( n \right)}=\frac{1}{2}\sum\limits_{i,j,k,l}{{\alpha _{ij}^{\left( n \right)}\alpha _{kl}^{\left( n \right)}\left\langle {{\mathbf{X}}_{ij}},{{\mathbf{X}}_{kl}} \right\rangle }}+C\sum\limits_{i,j}{\xi _{ij}^{\left( n \right)}} \\
 & +\frac{1}{2}\sum\limits_{i,j,k,l}{{\left( \beta _{ij}^{\left( n \right)}{{h}_{ij}}+\eta _{ij}^{\left( n \right)} \right)\left( \beta _{kl}^{\left( n \right)}{{h}_{kl}}+\eta _{kl}^{\left( n \right)} \right)\left\langle {{\mathbf{X}}_{ij}},{{\mathbf{X}}_{kl}} \right\rangle }}\\
 & -\sum\limits_{i,j}{\beta _{ij}^{\left( n \right)}},
\end{aligned}
\end{equation}
where $\alpha _{ij}^{\left( n \right)}$ and $\xi _{ij}^{\left( n \right)}$ are the feasible solutions to the primal problem, $\beta _{ij}^{\left( n \right)}$ and $\eta _{ij}^{\left( n \right)}$ are the feasible solutions to the dual problem, and $\text{DualGap}_{\text{NCML}}^{\left( n \right)}$ is the duality gap in the $n$th iteration. As $\eta _{ij}^{\left( n \right)}$ and $\mu _{ij}^{\left( n \right)}$ are the optimal solutions to the primal subproblem on $\boldsymbol{\eta }$ in (23) and its dual problem in (25), respectively, the duality gap of subproblem on $\boldsymbol{\eta }$ is zero, i.e.,
\begin{equation}
\begin{aligned}
& \frac{1}{2}\sum\nolimits_{i,j}{\sum\nolimits_{k,l}{\eta _{ij}^{\left( n \right)}\eta _{kl}^{\left( n \right)}\left\langle {{\mathbf{X}}_{ij}},{{\mathbf{X}}_{kl}} \right\rangle }}+\sum\nolimits_{i,j}{\eta _{ij}^{\left( n \right)}\gamma _{ij}^{\left( n \right)}}\\
& +\frac{1}{2}\sum\nolimits_{i,j}{\sum\nolimits_{k,l}{\mu _{ij}^{\left( n \right)}\mu _{kl}^{\left( n \right)}\left\langle {{\mathbf{X}}_{ij}},{{\mathbf{X}}_{kl}} \right\rangle }}-\sum\nolimits_{i,j}{\gamma _{ij}^{\left( n \right)}}\mu _{ij}^{\left( n \right)}=0.
\end{aligned}
\end{equation}
As shown in (26), $\alpha _{ij}^{\left( n \right)}$ and $\mu _{ij}^{\left( n \right)}$ should be equal. We substitute (28) into (27) as follows:
\begin{equation}
\begin{aligned}
\text{DualGap}_{\text{NCML}}^{\left( n \right)}=C\sum\nolimits_{i,j}{\xi _{ij}^{\left( n \right)}}-\sum\nolimits_{i,j}{\beta _{ij}^{\left( n \right)}}+\sum\nolimits_{i,j}{\mu _{ij}^{\left( n \right)}\gamma _{ij}^{\left( n \right)}}.
\end{aligned}
\end{equation}
\par
Based on the KKT conditions of the NCML dual problem in (19), $\xi _{ij}^{\left( n \right)}$ can be obtained by (\ref{xiijn}) (see page 7), where $\left[ z \right]=\max \left( z,0 \right)$ and $b^{(n)}$ can be obtained by\\
\begin{figure*}
\begin{equation}
\label{xiijn}
\xi _{ij}^{\left( n \right)}=\left\{ \begin{aligned}
  & 0\quad \text{for}\ \text{all}\ \beta _{ij}^{\left( n \right)}<C \\
 & {{\left[ 1-{{h}_{ij}}\left( \sum\nolimits_{k,l}{\alpha _{kl}^{\left( n \right)}\left\langle {{\mathbf{X}}_{ij}},{{\mathbf{X}}_{kl}} \right\rangle }+{{b}^{\left( n \right)}} \right) \right]}_{+}}={{\left[ \delta _{ij}^{\left( n+1 \right)}-{{h}_{ij}}\left( \gamma _{ij}^{\left( n \right)}+{{b}^{\left( n \right)}} \right) \right]}_{+}}\quad \text{for}\ \text{all}\ \beta _{ij}^{\left( n \right)}=C. \\
\end{aligned} \right.
\end{equation}
\end{figure*}
\begin{equation}
\label{bneq}
\begin{aligned}
{{b}^{\left( n \right)}} & =\frac{1}{{{h}_{ij}}}-\sum\nolimits_{k,l}{\alpha _{kl}^{\left( n \right)}\left\langle {{\mathbf{X}}_{ij}},{{\mathbf{X}}_{kl}} \right\rangle }\\
 & =\frac{\delta _{ij}^{\left( n+1 \right)}}{{{h}_{ij}}}-\gamma _{ij}^{\left( n \right)}\quad \text{for}\ \text{all}\ 0<\beta _{ij}^{\left( n \right)}<C.
\end{aligned}
\end{equation}
Please refer to \textbf{Appendix B} for the detailed derivation of $\xi _{ij}^{\left( n \right)}$ and ${{b}^{\left( n \right)}}$.
\par
\begin{figure}[ht]
\centering
\includegraphics[width=0.6\columnwidth]{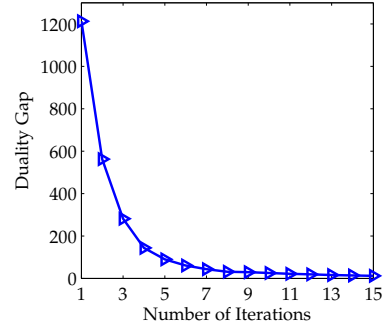}
\caption{Duality gap vs. number of iterations on the \emph{PenDigits} dataset for NCML.}
\label{Dualgap_NCML}
\end{figure}

Fig. \ref{Dualgap_NCML} plots the curve of duality gap versus the number of iterations on the \emph{PenDigits} dataset by NCML. One can see that the duality gap converges to zero in 15 iterations, and NCML reaches the global optimum. In the implementation of \textbf{Algorithm \ref{alg:NCML}}, we adopt the following termination condition:
\begin{equation}
\begin{aligned}
\text{DualGap}_{\text{NCML}}^{\left( t \right)}<\varepsilon \cdot \text{DualGap}_{\text{NCML}}^{\left( 1 \right)},
\end{aligned}
\end{equation}
where $\varepsilon $ is a small constant and we set $\varepsilon =0.01$ in the experiment.
\subsection{Remarks}
{\bf Computational complexity:} We use the same strategy as that in PCML to construct the pairwise constraints for NCML. In each iteration, NCML calls for the SVM solver twice while PCML calls for it only once. When the SMO-type algorithm \cite{Platt1999} is adopted for SVM training, the computational complexity of NCML is $O\left(k^2N^2d\right)$. One extra advantage of NCML lies in its lower computational cost with respect to $d$, which involves the computation of $\left\langle {{\mathbf{X}}_{ij}},{{\mathbf{X}}_{kl}} \right\rangle $ and the construction of matrix $\mathbf{M}$. Since $\left\langle {{\mathbf{X}}_{ij}},{{\mathbf{X}}_{kl}} \right\rangle ={{\left( {{({{\mathbf{x}}_{i}}-{{\mathbf{x}}_{j}})}^{T}}({{\mathbf{x}}_{k}}-{{\mathbf{x}}_{l}}) \right)}^{2}}$, the cost of computing $\left\langle {{\mathbf{X}}_{ij}},{{\mathbf{X}}_{kl}} \right\rangle $ is $O\left(d\right)$. The cost of constructing the matrix $\mathbf{M}$ is less than $O\left(kNd^2\right)$, and this operation is required only once after the convergence of $\boldsymbol{\beta}$ and $\boldsymbol{\mu}$.
\par
{\bf Nonlinear extensions:} Note that $\left\langle {{\mathbf{X}}_{ij}},{{\mathbf{X}}_{kl}} \right\rangle = \operatorname{tr}\left( \mathbf{X}_{ij}^{T}{{\mathbf{X}}_{kl}} \right)$ can be treated as an inner product of two pairs of samples: $\left(\mathbf{x}_i, \mathbf{x}_j\right)$ and $\left(\mathbf{x}_k, \mathbf{x}_l\right)$. Analogous to PCML, if we can define a kernel $K\left( ({{\mathbf{x}}_{i}},{{\mathbf{x}}_{j}}),({{\mathbf{x}}_{k}},{{\mathbf{x}}_{l}}) \right)$ on $\left(\mathbf{x}_i, \mathbf{x}_j\right)$ and $\left(\mathbf{x}_k, \mathbf{x}_l\right)$, we can substitute $\left\langle {{\mathbf{X}}_{ij}},{{\mathbf{X}}_{kl}} \right\rangle $ with $K\left( ({{\mathbf{x}}_{i}},{{\mathbf{x}}_{j}}),({{\mathbf{x}}_{k}},{{\mathbf{x}}_{l}}) \right)$ to develop new linear or even nonlinear metric learning algorithms, and the Mahalanobis distance between any two samples $\mathbf{x}_m$ and $\mathbf{x}_n$ can be formulated as:
\begin{equation}
\begin{aligned}
& {{\left( {{\mathbf{x}}_{m}}-{{\mathbf{x}}_{n}} \right)}^{T}}\mathbf{M}\left( {{\mathbf{x}}_{m}}-{{\mathbf{x}}_{n}} \right)=\\
& \sum\nolimits_{i,j}{{{\alpha }_{ij}}K\left( \left( {{\mathbf{x}}_{i}},{{\mathbf{x}}_{j}} \right),\left( {{\mathbf{x}}_{m}},{{\mathbf{x}}_{n}} \right) \right).}
\end{aligned}
\end{equation}
\par
 Another nonlinear extension strategy is to define a kernel $k\left( {{\mathbf{x}}_{i}},{{\mathbf{x}}_{j}} \right)$ on $\mathbf{x}_i$ and $\mathbf{x}_j$. Since $\left\langle {{\mathbf{X}}_{ij}},{{\mathbf{X}}_{kl}} \right\rangle ={{\left(\mathbf{x}_{i}^{T}{{\mathbf{x}}_{k}}-\mathbf{x}_{i}^{T}{{\mathbf{x}}_{l}}-\mathbf{x}_{j}^{T}{{\mathbf{x}}_{k}}+\mathbf{x}_{j}^{T}{{\mathbf{x}}_{l}} \right)}^{2}}$, we can substitute $\left\langle {{\mathbf{X}}_{ij}},{{\mathbf{X}}_{kl}} \right\rangle $ with ${{\left( k\left( {{\mathbf{x}}_{i}},{{\mathbf{x}}_{k}} \right)-k\left( {{\mathbf{x}}_{i}},{{\mathbf{x}}_{l}} \right)-k\left( {{\mathbf{x}}_{j}},{{\mathbf{x}}_{k}} \right)+k\left( {{\mathbf{x}}_{j}},{{\mathbf{x}}_{l}} \right) \right)}^{2}}$ and formulate the Mahalanobis distance between $\mathbf{x}_m$ and $\mathbf{x}_n$ as:
 \begin{equation}
\begin{aligned}
& {{\left( {{\mathbf{x}}_{m}}-{{\mathbf{x}}_{n}} \right)}^{T}}\mathbf{M}\left( {{\mathbf{x}}_{m}}-{{\mathbf{x}}_{n}} \right)\\
& =\sum\nolimits_{i,j}{{{\alpha }_{ij}}{{\left(
\begin{aligned}
& k\left( {{\mathbf{x}}_{i}},{{\mathbf{x}}_{m}} \right)-k\left( {{\mathbf{x}}_{i}},{{\mathbf{x}}_{n}} \right)\\
& -k\left( {{\mathbf{x}}_{j}},{{\mathbf{x}}_{m}} \right)+k\left( {{\mathbf{x}}_{j}},{{\mathbf{x}}_{n}} \right)
\end{aligned}
\right)}^{2}}}.
\end{aligned}
\end{equation}
 That is to say, NCML allows us to learn nonlinear metrics for histograms and structural data by designing proper kernel functions and incorporating appropriate regularizations on $\boldsymbol{\alpha}$. Metric learning for structural data beyond vector data has been recently receiving considerable research interests \cite{huang2011generalized, lim2013robust}, and NCML can provide a new perspective on this topic.
\par
{\bf SVM solvers:} Although our implementation is based on LibSVM, there are a number of well-studied SVM training algorithms, e.g., core vector machines \cite{tsang2005core}, LaRank \cite{bordes2007}, BMRM \cite{teo2007scalable}, and Pegasos \cite{shalev2011pegasos}, which can be utilized for large scale metric learning. Moreover, we can refer to the progresses in kernel methods \cite{belkin2006manifold,andrews2002support,evgeniou2004regularized} for developing semi-supervised, multiple instance, and multitask metric learning approaches.
\section{Experimental Results}
\label{experimentalresults}
We evaluate the proposed PCML and NCML models for $k$-NN classification ($k=1$) using 9 UCI datasets, 4 handwritten digit datasets, 2 face verification datasets and 2 person re-identification datasets. We compare PCML and NCML with the baseline Euclidean distance metric and 7 state-of-the-art metric learning models, including NCA \cite{goldberger2004NCA}, ITML \cite{davis2007information}, MCML \cite{globerson2005MCML}, LDML \cite{guillaumin2009LDML}, LMNN \cite{weinberger2009LMNNJMLR}, PLML \cite{wang2012PLML}, and DML-eig \cite{ying2012distance}. On each dataset, if the partition of training set and test set is not defined, we evaluate the performance of each method by 10-fold cross-validation, and the classification error rate and training time are obtained by averaging over 10 runs of 10-fold cross-validation. PCML and NCML are implemented using the LibSVM\footnote{\url{http://www.csie.ntu.edu.tw/~cjlin/libsvm/}} toolbox. The source codes of NCA\footnote{\url{http://www.cs.berkeley.edu/~fowlkes/software/nca/}}, ITML\footnote{\url{http://www.cs.utexas.edu/~pjain/itml/}}, MCML\footnote{\url{http://homepage.tudelft.nl/19j49/Matlab_Toolbox_for_Dimensionality_Reduction.html}}, LDML\footnote{\url{http://lear.inrialpes.fr/people/guillaumin/code.php}}, LMNN\footnote{\url{http://www.cse.wustl.edu/~kilian/code/code.html}}, PLML\footnote{\url{http://cui.unige.ch/~wangjun/}}, and DML-eig\footnote{\url{http://empslocal.ex.ac.uk/people/staff/yy267/software.html}} are online available, and we tune their parameters to get the best results.
\subsection{Results on the UCI Datasets}
We first use 9 datasets from the UCI Machine Learning Repository \cite{frank2010uci} to evaluate the proposed models. The information of the 9 UCI datasets is summarized in Table \ref{UCI_description}. On the \emph{Satellite}, \emph{SPECTF Heart}, and \emph{Letter} datasets, the training set and test set are defined. On the other datasets, we use 10-fold  cross-validation to evaluate the metric learning models.
\begin{table}[!t]
\renewcommand{\arraystretch}{1.3}
\caption{The UCI datasets used in our experiments.}
\label{UCI_description}
\centering
\scriptsize
\begin{tabular}{m{1.8cm}m{1.6cm}m{1.0cm}m{1.3cm}m{0.8cm}}
\hline
\bfseries Dataset & \bfseries \begin{flushleft}\# of training samples\end{flushleft} & \bfseries \begin{flushleft}\# of test samples\end{flushleft} & \bfseries \begin{flushleft}Feature dimension\end{flushleft} & \bfseries \begin{flushleft}\# of classes\end{flushleft} \\
\hline
Breast Tissue & 96 & 10 & 9 & 6\\
Cardiotocography & 1,914 & 212 & 21 & 10\\
ILPD & 525 & 58 & 10 & 2\\
Letter & 16,000 & 4,000 & 16 & 26\\
Parkinsons & 176 & 19 & 22 & 2\\
Satellite & 4,435 & 2,000 & 36 & 6\\
Segmentation & 2,079 & 231 & 19 & 7\\
Sonar & 188 & 20 & 60 & 2\\
SPECTF Heart & 80 & 187 & 44 & 2\\
\hline
\end{tabular}
\vskip -0in
\vspace{-5mm}
\end{table}
\par
\begin{figure}[ht]
\centering
\subfigure[]{
\includegraphics[width=0.46\columnwidth]{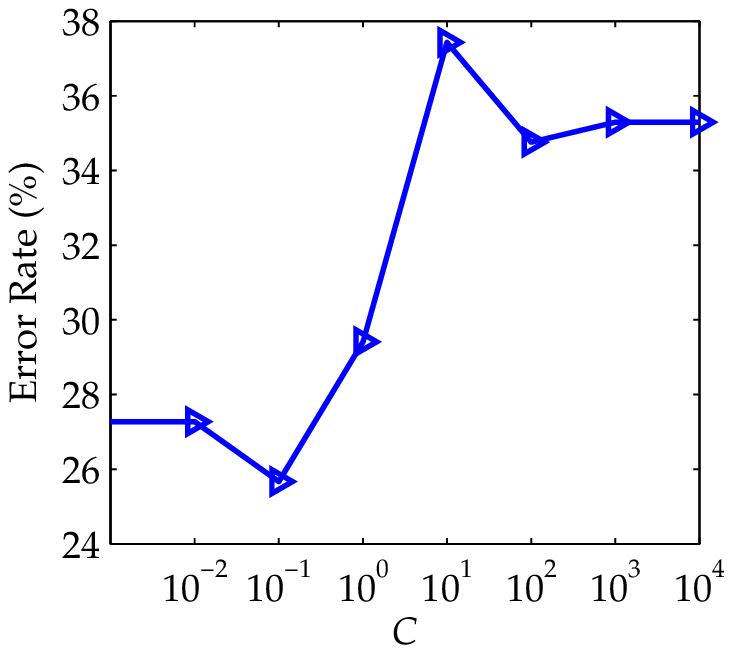}
}
\subfigure[]{
\includegraphics[width=0.46\columnwidth]{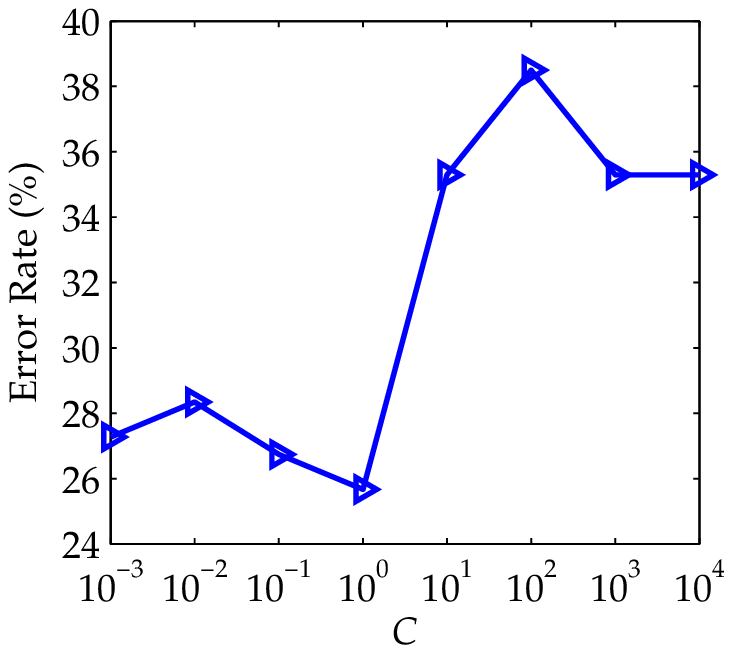}
}
\caption{Classification error rate (\%) versus $C$. (a) PCML; (b) NCML.}
\label{C_param}
\end{figure}

The proposed PCML and NCML methods involve only one hyper-parameter, i.e., the regularization parameter $C$. We simply adopt the cross-validation strategy to select $C$ by investigating the influence of $C$ on the classification error rate. Fig. \ref{C_param} shows the curves of classification error rate versus $C$ for PCML and NCML on the \emph{SPECTF Heart} dataset. The curves on other datasets are similar. We can observe that when $C < 1$, the classification error rates of PCML and NCML will be low and stable. When $C$ is higher than $1$, the classification error rates jump dramatically. Thus, we set $C<1$ in our experiments.
\par
We compare the classification error rates of the competing methods in Table \ref{UCI_errorrate}. On the \emph{Cardiotocography} and \emph{Segmentation} datasets, PCML achieves the lowest error rates. On the \emph{Segmentation} and \emph{SPECTF Heart} datasets, NCML achieves the lowest error rates. The average ranks of competing methods are listed in the last row of Table \ref{UCI_errorrate}. On each dataset, we rank the methods based on their error rates, i.e., we assign rank 1 to the method with the lowest error rate and rank 2 to the method with the second lowest error rate, and so on. The average rank is defined as the mean rank of one method over the nine datasets, which can provide a fair comparison of the learning methods \cite{demvsar2006statistical}. From Table \ref{UCI_errorrate}, we can see that both PCML and NCML achieve the first and second best average ranks, respectively, demonstrating strong classification capability for general classification tasks.

\begin{table*}[!t]
\renewcommand{\arraystretch}{1.3}
\caption{Classification error rate (\%) on the UCI datasets.}
\label{UCI_errorrate}
\begin{center}
%\centering
\begin{tabular}{ccccccccccc}
\hline
\bfseries Dataset & \bfseries Euclidean & \bfseries NCA & \bfseries ITML & \bfseries MCML & \bfseries LDML & \bfseries LMNN & \bfseries PLML & \bfseries DML-eig & \bfseries PCML & \bfseries NCML \\
\hline
Breast Tissue & \bfseries 31.00 & 41.27 & 35.82 & 32.09 & 48.00 & 34.37 & 34.13 & 33.13 & 38.00 & 35.37 \\
Cardiotocography & 21.40 & 21.16 & 18.67 & 22.29 & 22.26 & 19.21 & 18.54 & 29.31 & \bfseries 18.50 & 18.69 \\
ILPD & 35.69 & 34.65 & 35.35 & 35.49 & 35.84 & 34.12 & \bfseries 31.61 & 36.87 & 33.96 & 32.43 \\
Letter & 4.33 & \bfseries 2.47 & 3.80 & 4.20 & 11.05 & 3.45 & 3.28 & 3.85 & 2.67 & 2.72 \\
Parkinsons & \bfseries 4.08 & 6.63 & 6.13 & 9.84 & 7.15 & 5.26 & 8.84 & 7.82 & 5.68 & 7.26 \\
Satellite & 10.95 & 10.40 & 11.45 & 15.65 & 15.90 & \bfseries 10.05 & 11.85 & 10.90 & 11.15 & 11.10 \\
Segmentation & 2.86 & 2.51 & 2.73 & 2.60 & 2.86 & 2.64 & 2.68 & 2.97 & \bfseries 2.12 & \bfseries 2.12 \\
Sonar & 12.98 & 15.40 & 12.07 & 24.29 & 22.86 & \bfseries 11.57 & 12.07 & 15.07 & 12.71 & 13.29 \\
SPECTF Heart & 38.50 & 26.74 & 34.76 & 38.50 & 33.16 & 34.76 & 27.27 & 31.02 & 28.88 & \bfseries 25.67 \\
\hline
\emph{Average Rank} & \emph{5.78} & \emph{4.56} & \emph{5.44} & \emph{7.56} & \emph{8.44} & \emph{4.00} & \emph{4.33} & \emph{7.00} & \bfseries \emph{3.56} & \emph{3.89}\\
\hline
\end{tabular}
\end{center}
\vskip -0in
\vspace{-5mm}
\end{table*}

\par
\begin{figure}[ht]
\vskip 0in
\begin{center}
\centerline{\includegraphics[width=1.0\columnwidth]{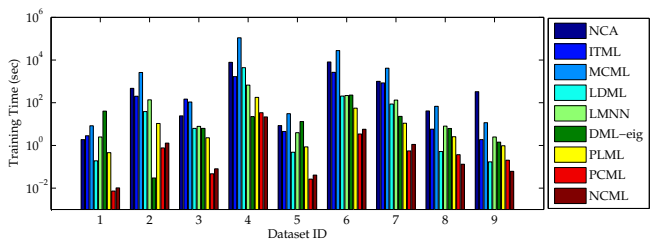}}
\caption{Training time (s) of NCA, ITML, MCML, LDML, LMNN, DML-eig, PLML, PCML and NCML. From 1 to 9, the Dataset ID represents \emph{Breast Tissue}, \emph{Cardiotocography}, \emph{ILPD}, \emph{Letter}, \emph{Parkinsons}, \emph{Satellite}, \emph{Segmentation}, \emph{Sonar} and \emph{SPECTF Heart}.}
\label{UCI_TrainingTime}
\end{center}
\vskip -0.2in
\vspace{0mm}
\end{figure}

We then compare the training time of competing metric learning methods in Fig. \ref{UCI_TrainingTime}. All the experiments are run in a PC with 4 Intel Core i5-2410 CPUs (2.30 GHz) and 16GB RAM. Clearly, the proposed PCML and NCML are the fastest in most cases. Although DML-eig is faster than PCML on the \emph{Letter} dataset, its classification error rate on this dataset is much higher than PCML and NCML. On average, PCML and NCML are 23 and 18 times faster than PLML, the third fastest algorithm, respectively.
\subsection{Handwritten Digit Recognition}
We further evaluate the proposed methods on four handwritten digit datasets: \emph{MNIST}, \emph{Pen-based recognition of handwritten Digits data set (PenDigits)}, \emph{Semeion} and \emph{USPS}. Table \ref{Digits_description} summarizes the basic information of these four handwritten digit datasets. On the \emph{MNIST}, \emph{PenDigits}, and \emph{USPS} datasets, we use the defined training sets to train the metrics, and use the defined test sets to compute the classification error rates. On the \emph{Semeion} dataset, we use 10-fold cross-validation to evaluate the metric learning methods, and the classification error rate and training time are obtained by averaging over 10 runs of 10-fold cross-validation.
\begin{table}[!t]
\renewcommand{\arraystretch}{1.3}
\caption{The handwritten digit datasets used in the experiments.}
\label{Digits_description}
\begin{center}
\centering\arraybackslash
\scriptsize
\begin{tabular}{m{0.9cm}m{1.6cm}m{1.1cm}m{1cm}m{1cm}m{0.8cm}}
\hline
\bfseries Dataset & \bfseries \begin{flushleft}\# of training samples\end{flushleft} & \bfseries \begin{flushleft}\# of test samples\end{flushleft} & \bfseries dimension & \bfseries \begin{flushleft}PCA dimension\end{flushleft} & \bfseries \begin{flushleft}\# of classes\end{flushleft} \\
\hline
MNIST & 60,000 & 10,000 & 784 & 100 & 10\\
PenDigits & 7,494 & 3,498 & 16 & N/A & 10\\
Semeion & 1,434 & 159 & 256 & 100 & 10\\
USPS & 7,291 & 2,007 & 256 & 100 & 10\\
\hline
\end{tabular}
\end{center}
\end{table}
\par
As the dimensions of images in the \emph{MNIST}, \emph{Semeion} and \emph{USPS} datasets are relatively high, we use principal component analysis (PCA) to reduce the feature dimension to 100, and train the metrics in the PCA subspace. Table \ref{Digits_errorrate} lists the classification error rates of the ten competing methods on the four handwritten digit datasets. The last row of Table \ref{Digits_errorrate} lists the average ranks of the competing methods. We do not report the error rate and training time of MCML on the \emph{MNIST} dataset because MCML requires too large memory space (more than 30 GB) on this dataset and cannot run in our PC. From Table \ref{Digits_errorrate}, we can see that both PCML and NCML achieve the best average rank. Again, the results indicate that the proposed methods have better classification performance.
\begin{table*}[!t]
\renewcommand{\arraystretch}{1.3}
\caption{Comparison of classification error rate (\%) on the handwritten digit datasets.}
\label{Digits_errorrate}
\vskip 0in
\begin{center}
\begin{tabular}{ccccccccccc}
\hline
\bfseries Dataset & \bfseries Euclidean & \bfseries NCA & \bfseries ITML & \bfseries MCML & \bfseries LDML & \bfseries LMNN & \bfseries DML-eig & \bfseries PLML & \bfseries PCML & \bfseries NCML \\
\hline
MNIST & 2.87 & 5.46 & 2.89 & N/A & 6.05 & \bfseries 2.28 & 5.06 & 2.54 & 3.85 & 2.80 \\
PenDigits & 2.26 & 2.23 & 2.29 & 2.26 & 6.20 & 2.52 & 3.75 & 2.46 & \bfseries 2.06 & \bfseries 2.06 \\
Semeion & 8.54 & 8.60 & 5.71 & 11.23 & 11.98 & 6.09 & 5.72 & 7.66 & \bfseries 4.83 & 5.53 \\
USPS & \bfseries 5.08 & 5.68 & 6.33 & \bfseries 5.08 & 8.77 & 5.38 & 11.36 & 6.73 & 5.33 & 5.43 \\
\hline
\emph{Average Rank} & \emph{4.00} & \emph{6.25} & \emph{5.25} & \emph{4.67} & \emph{9.50} & \emph{4.50} & \emph{7.50} & \emph{5.75} & \bfseries \emph{2.75} & \bfseries \emph{2.75} \\
\hline
\end{tabular}
\end{center}
\vskip -0in
\vspace{-5mm}
\end{table*}
\par
\begin{figure}[ht]
\vskip 0in
\begin{center}
\centerline{\includegraphics[width=1.0\columnwidth]{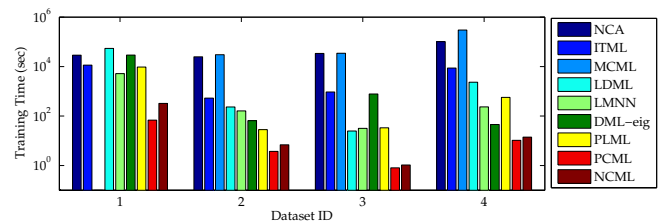}}
\caption{Training time (s) of NCA, ITML, MCML, LDML, LMNN, DML-eig, PLML, PCML and NCML. From 1 to 4, the Dataset ID represents \emph{MNIST}, \emph{PenDigits}, \emph{Semeion} and \emph{USPS}.}
\label{Digits_TrainingTime}
\end{center}
\vskip -0in
\vspace{-5mm}
\end{figure}

 All the experiments were executed in the same PC as used in Subsection 5.1. Fig. \ref{Digits_TrainingTime} compares the training time of NCA, ITML, MCML, LDML, LMNN, DML-eig, PLML, PCML, and NCML. Clearly, the proposed PCML and NCML methods are much faster than the other methods. On average, PCML and NCML are 61 and 27 times faster than PLML, the third fastest algorithm, respectively. One can conclude that PCML and NCML offer promising solutions to effective and efficient metric learning.
\par
\begin{figure}[ht]
\vskip 0in
\begin{center}
\centerline{\includegraphics[width=0.7\columnwidth]{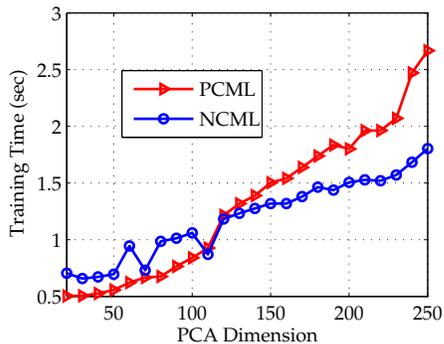}}
\caption{Training time (s) vs. PCA dimension on the \emph{Semeion} dataset.}
\label{TrainTimePCA_Semeion}
\end{center}
\vskip -0in
\vspace{-5mm}
\end{figure}

Finally, we compare the running time of PCML and NCML under different feature dimensions $d$. As analyzed in Subsections 3.4 and 4.4, the time complexities of PCML and NCML are $O(N^2d + d^3)$ and $O(N^2d)$, respectively. Fig. \ref{TrainTimePCA_Semeion} shows the training time on the \emph{Semeion} dataset with different PCA dimensions. We can see that when the dimension is lower than 110, the training time of NCML is longer than PCML. When the dimension is higher than 110, the training time of PCML increases and becomes longer than NCML.
\subsection{Face Verification}
In this subsection, we evaluate the proposed methods for face verification using two challenging face databases: Labeled Faces in the Wild (LFW) \cite{huang2007LFW} and Public Figures (PubFig) \cite{kumar2009attribute}.
\subsubsection{The LFW Database}
The face images in the LFW database were collected from the Internet and demonstrate large variations of pose, illumination, expression, etc. The database consists of 13,233 face images from 5,749 persons. Under the image restricted setting, the performance of a face verification method is evaluated by 10-fold cross validation. For each of the 10 runs, the database provides 300 positive pairs and 300 negative pairs for testing, and 5,400 image pairs for training. The verification rate and Receiver Operator Characteristic (ROC) curve of each method are obtained by averaging over the 10 runs.
\par
In our experiments, we use the SIFT \cite{lowe2004distinctive} features and the attribute features provided by \cite{guillaumin2009LDML} and \cite{kumar2009attribute} to evaluate the metric learning methods. Since the dimension of SIFT features is high (i.e., 128 $\times$ 3 $\times$ 9), PCA is used to reduce the feature dimension to 150. Under the restricted setting of the LFW database, we only know whether two images are matched or not for the given pairs. In the training stage, we use the training pairs to train a Mahalanobis distance metric. In the test stage, we compare the Mahalanobis distance of the test pair with a threshold $t$ to decide whether the two images are matched or not.
%\begin{figure}[ht]
%\vskip 0in
%\begin{center}
%\centerline{\includegraphics[width=0.7\columnwidth]{LFW-image}}
%\caption{The example pairs on the LFW dataset.}
%\label{LFW_image}
%\end{center}
%\vskip -0in
%\vspace{-5mm}
%\end{figure}
\par
\begin{figure*}[ht]
\centering
\subfigure[SIFT]{
\includegraphics[width=0.6\columnwidth]{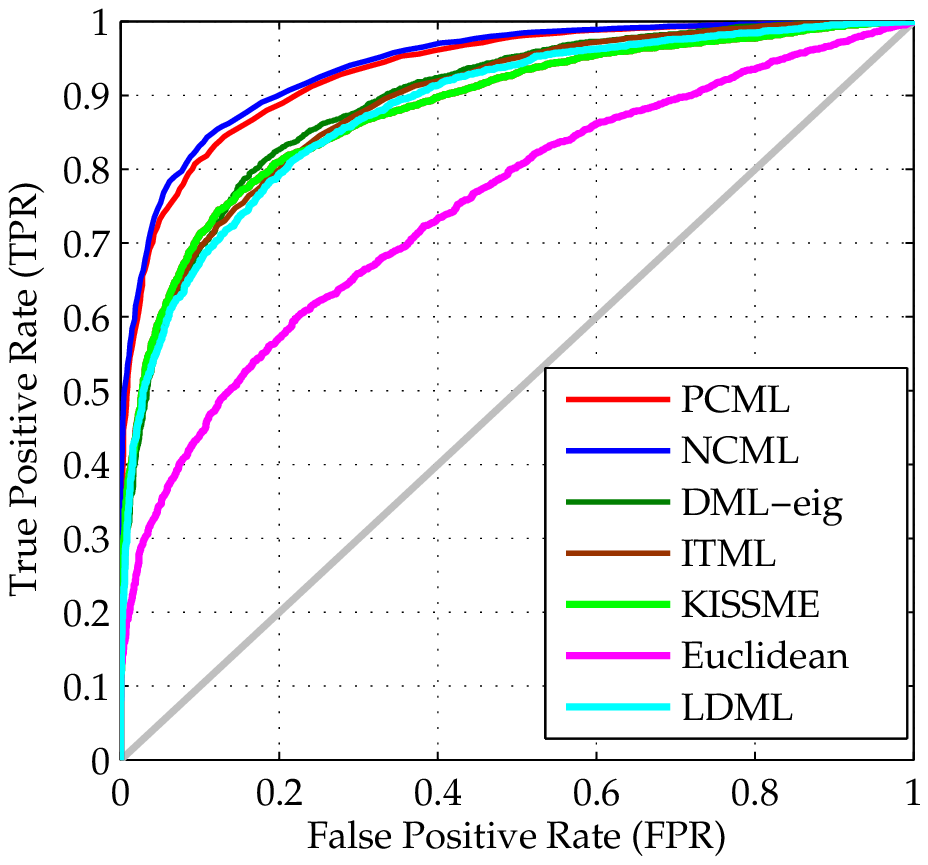}
}
\subfigure[Attribute]{
\includegraphics[width=0.6\columnwidth]{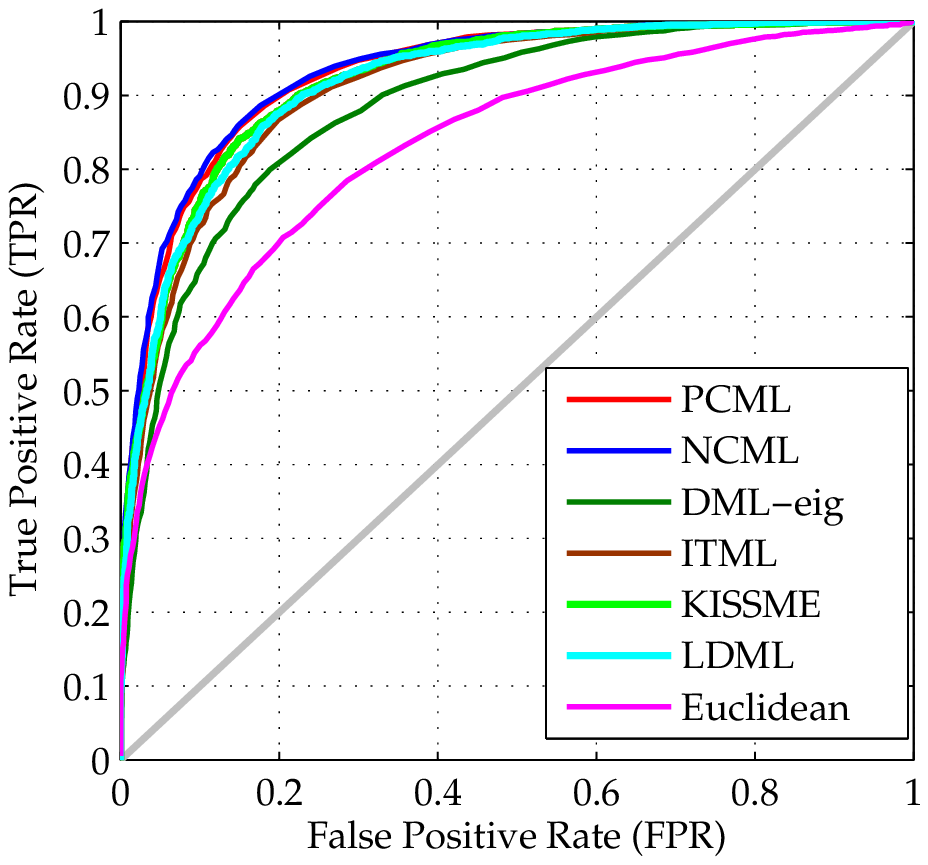}
}
\subfigure[SIFT + Attribute]{
\includegraphics[width=0.6\columnwidth]{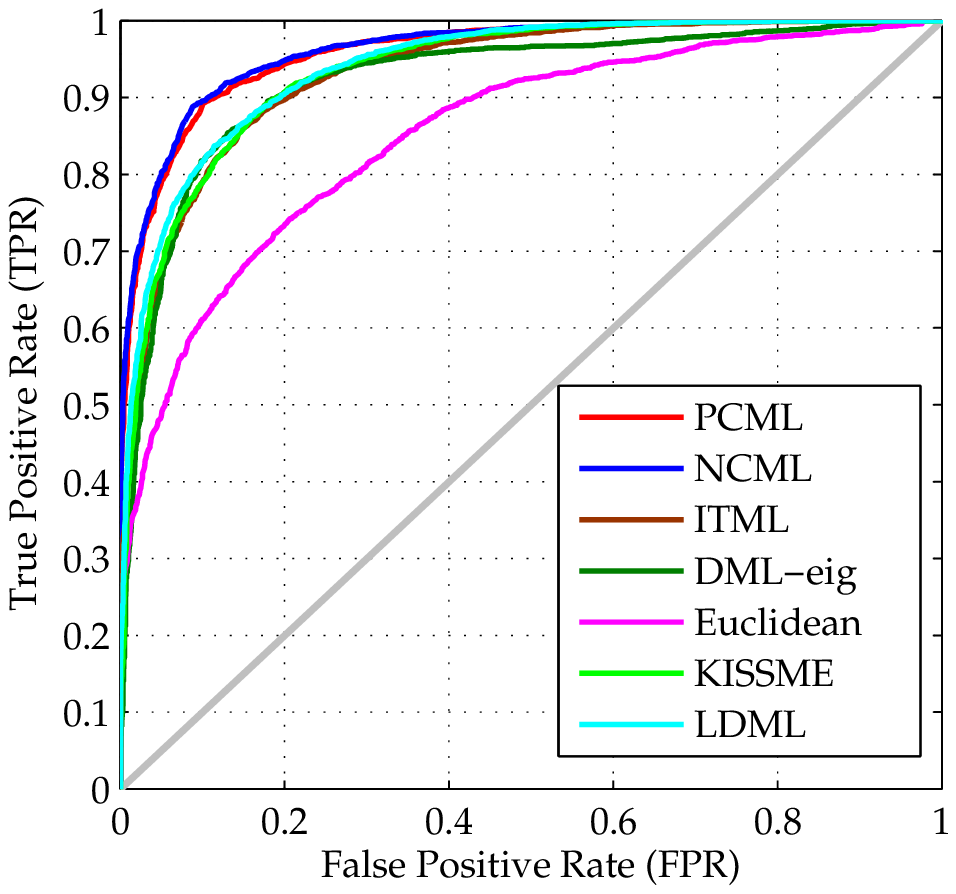}
}
\caption{The ROC curves of different metric learning methods on the LFW-funneled dataset under the image restricted setting\cite{ying2012distance,kostinger2012large,guillaumin2009LDML}. (a) SIFT feature; (b) Attribute feature; (c) SIFT + Attribute feature.}
\label{LFW_ROC}
%\vskip -0in
%\vspace{-5mm}
\end{figure*}

We report the ROC curves of PCML, NCML, DML-eig \cite{ying2012distance}, ITML \cite{davis2007information}, KISSME \cite{kostinger2012large}, LDML \cite{guillaumin2009LDML} and Euclidean distance in Fig. \ref{LFW_ROC}. We also compare the verification accuracies of PCML and NCML and other metric learning methods by using the SIFT  and the attribute features in Table \ref{LFW_RATE}. It can be seen that the proposed PCML and NCML methods perform much better than all the other competing methods. Using the combination of SIFT and Attribute features, the verification accuracies of PCML (\textbf{89.00\%}) and NCML (\textbf{89.50\%}) are higher than the third best method, i.e. DML-eig (85.65\%), by 3.35\% and 3.85\%, respectively. We also compare the training time of the competing methods in Table \ref{LFW_RATE}. The training time of PCML and NCML is shorter than the other methods except for KISSME. The reason is that KISSME is a one-pass training approach. Although KISSME is faster, its verification accuracy is much lower than PCML and NCML.

\begin{table}[!t]
\renewcommand{\arraystretch}{1.3}
\caption{Verification accuracies (\%) and training time (s) of competing metric learning methods on the LFW-funneled dataset under the image restricted setting.}
\label{LFW_RATE}
\vskip 0in
\begin{center}
%\begin{tabular}
\begin{tabular}{m{1.2cm}|m{0.8cm}m{1cm}m{1.5cm}|m{0.8cm}m{1cm}}
\hline
\multirow{2}{*}{\bfseries Method} & \multicolumn{3}{c|}{\bfseries Verification Accuracy (\%)} & \multicolumn{2}{c}{\bfseries Training Time (s)}\\
\cline{2-6}
& \bfseries SIFT & \bfseries Attribute & \bfseries SIFT + Attribute & \bfseries SIFT & \bfseries Attribute\\
\hline
PCML & 85.70 & 84.70 & 89.00& 13.22 & 14.17\\
NCML & \textbf{86.45} & \textbf{85.45} & \textbf{89.50} & 31.62 & 27.55\\
DML-eig\cite{ying2012distance} & 81.27 & 80.13 & 85.65 & 1931.50 & 113.79\\
ITML\cite{kostinger2012large} & 82.40 & 82.98 & 85.50 & 3341.80 & 3222.40\\
LDML\cite{guillaumin2009LDML} & 79.27 & 83.40 & 86.02 & 1316.60 & 543.08\\
KISSME\cite{kostinger2012large} & 80.50 & 84.60 & 85.39 & 0.22 & 0.05\\
Euclidean & 68.10 & 75.25 & 76.53 & 0 & 0\\
\hline
\end{tabular}
\end{center}
\end{table}
\subsubsection{The PubFig Database}
The PubFig database \cite{kumar2009attribute} contains 58,797 face images of 200 persons with large variations in pose, lighting, expression, scene, camera, imaging conditions and parameters, etc. In this database, the face verification methods are also evaluated using 10-fold cross validation. Among the given 20,000 image pairs, we randomly select 18,000 pairs for training and use the remaining 2,000 pairs for testing in each run. The ROC curves and verification rates are obtained by averaging over the 10 runs.
\par
\begin{figure}[ht]
\vskip 0in
\begin{center}
\centerline{\includegraphics[width=0.7\columnwidth]{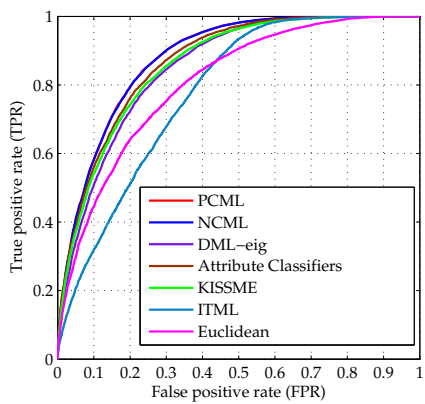}}
\caption{The ROC curves of different methods on the PubFig database (the curves of PCML and NCML almost coincide).}
\label{PubFig_ROC}
\end{center}
\vskip -0in
\vspace{-5mm}
\end{figure}

We use the attribute features provided by \cite{kumar2009attribute} to evaluate the competing methods. Fig. \ref{PubFig_ROC} shows the ROC curves of PCML, NCML, KISSME \cite{kostinger2012large}, ITML \cite{davis2007information}, DML-eig \cite{ying2012distance}, Attribute Classifiers \cite{kumar2009attribute} and the baseline Euclidean distance. It can be seen that the performance of PCML and NCML is similar, and is superior to that of the other methods.
\par
We further report the verification rates of PCML, NCML and the other methods in Table \ref{PubFig_Rate}. One can see that PCML and NCML perform better than the other methods. The accuracies of PCML (\textbf{79.71\%}) and NCML (\textbf{79.75\%}) are higher than the third best method, i.e., Attribute Classifiers (78.65\%), by 1.06\% and 1.10\%, respectively. The training time of PCML, NCML and other metric learning methods is also listed in Table \ref{PubFig_Rate}. It can be seen that PCML and NCML are much faster than ITML and DML-eig.

\par

\begin{table}[!t]
\renewcommand{\arraystretch}{1.3}
\caption{Verification accuracies (\%) and training time (s) of competing methods on the PubFig database.}
\label{PubFig_Rate}
\vskip 0in
\begin{center}
\begin{tabular}{m{2.7cm}m{2.3cm}m{2cm}}
\hline
\bfseries Methods & \bfseries \begin{flushleft}Verification Accuracy (\%)\end{flushleft} & \bfseries \begin{flushleft}Training Time (s)\end{flushleft}\\
\hline
PCML & 79.71 & 118.55\\
NCML & \textbf{79.75} & 216.38\\
KISSME\cite{kostinger2012large} & 77.60 & 0.09\\
ITML\cite{kostinger2012large} & 69.30 & 3796.50\\
Attribute Classifiers\cite{kumar2009attribute} & 78.65 & -\\
DML-eig\cite{ying2012distance} & 77.36 & 1132.30\\
Euclidean & 72.50 & 0\\
\hline
\end{tabular}
\end{center}
\end{table}

\subsection{Person Re-identification}
In this subsection, we evaluate the performance of the proposed methods for person re-identification, i.e., recognizing a person at different locations and at different times \cite{gong2014person}. Two challenging person re-identification databases, the Viewpoint Invariant Pedestrian Recognition (VIPeR) database \cite{gray2008viewpoint} and the Context Aware Vision using Image-based Active Recognition for Re-Identification (CAVIAR4REID) database \cite{cheng2011custom} are used to assess the performance of the proposed methods.

\subsubsection{The VIPeR Database}
The VIPeR database contains 1,264 pedestrian images of 632 persons from two camera viewspoints (camera A and camera B). For each person, there are two images taken from different viewpoints with a change of 90 degrees. In our experiments, we randomly select 316 persons and use their images for training, and use the images of the other 316 persons for testing. For the testing images, we use the images taken by camera B as the probe set and the images from camera A as the gallery set. Finally, 10 partitions of training and test sets are constructed, and the average accuracy over the 10 test sets is computed as the final accuracy.
%\begin{figure}[ht]
%\vskip 0in
%\begin{center}
%\centerline{\includegraphics[width=0.7\columnwidth]{VIPeR}}
%\caption{The example images in VIPeR dataset.}
%\label{VIPeR_image}
%\end{center}
%\vskip -0in
%\vspace{-5mm}
%\end{figure}

\begin{table}[!t]
\renewcommand{\arraystretch}{1.3}
\caption{Person re-identification accuracies (\%) and training time (s) on the VIPeR dataset.}
\label{VIPeR_Rate}
\vskip 0in
\begin{center}
\begin{tabular}{p{1.2cm}|m{0.6cm}m{0.6cm}m{0.6cm}m{0.6cm}m{0.6cm}|m{1.2cm}}
\hline
\multirow{2}{*}{Methods} & \multicolumn{5}{c|}{\bfseries Accuracy (\%)} & \multirow{2}{1.2cm}{\bfseries Training\\Time\\(s)}\\
\cline{2-6}
& \bfseries Rank 1 & \bfseries Rank 25 & \bfseries Rank 50 & \bfseries Rank 80 & \bfseries Rank 100\\
\hline
PCML & 19.40 & 80.60 & \textbf{93.77} & \textbf{97.25} & 98.23 & 4.94\\
NCML & \textbf{21.04} & \textbf{82.28} & 93.07 & \textbf{97.25} & \textbf{98.32} & 9.05\\
KISSME\cite{kostinger2012large} & 19.60 & 80.70 & 91.80 & 96.68 & 97.78 & 0.07\\
LMNN\cite{kostinger2012large} & 16.61 & 72.94 & 88.13 & 94.30 & 96.36 & 437.43\\
ITML\cite{kostinger2012large} & 15.66 & 74.21 & 88.29 & 95.41 & 96.99 & 1199.10\\
DML-eig\cite{ying2012distance} & 8.07 & 50.47 & 65.82 & 77.69 & 82.44 & 47.03\\
Euclidean & 10.90 & 44.94 & 60.76 & 70.09 & 74.37 & 0\\
\hline
\end{tabular}
\end{center}
\end{table}
%\par
%For each images in one viewpoint, we can select the image of the same person from another viewpoint to construct the positive pair, and randomly select an image of different persons from another viewpoint to construct the negative pair. Following this strategy, we can construct 316 positive pairs and 316 negative pairs for each run.
\par
\begin{figure}[ht]
\vskip 0in
\begin{center}
\centerline{\includegraphics[width=0.6\columnwidth]{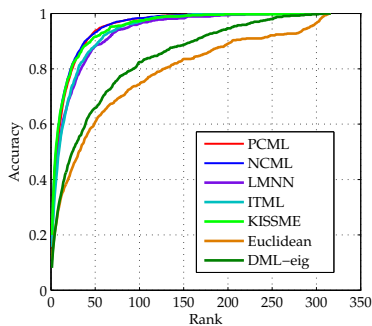}}
\caption{The CMC curves on the VIPeR dataset.}
\label{VIPER_CMC}
\end{center}
\vskip -0in
\vspace{-5mm}
\end{figure}

We report the Cumulative Matching Characteristic (CMC) curves of the competing methods in Fig. \ref{VIPER_CMC}. We also compare their accuracies under different ranks in Table \ref{VIPeR_Rate}. From Fig. \ref{VIPER_CMC} and Table \ref{VIPeR_Rate}, one can see that both PCML and NCML outperform LMNN, ITML and Euclidean distance significantly under all ranks. When the rank is no more than 25, PCML performs similarly to KISSME, while NCML outperforms KISSME. When the rank is between 25 and 200, both PCML and NCML perform better than KISSME. The training time of the metric learning methods is also reported in Table \ref{VIPeR_Rate}. We can see that both PCML and NCML are much more efficient than LMNN and ITML in training.

\subsubsection{The CAVIAR4REID Database}
CAVIAR4REID consists of 1,220 pedestrian images from 72 persons, where the images are extracted from the shopping center scenario of the CAVIAR database \cite{cheng2011custom}. The database covers a large range of image resolution and pose variation. The minimum and maximum image sizes in the CAVIAR4REID database are $17\times39$ and $72\times144$, respectively. Following \cite{zhou2009hierarchical} and \cite{li2013learning}, we use the hierarchical Gaussian (HG) features to evaluate the metric learning methods.
%\begin{figure}[ht]
%\vskip 0in
%\begin{center}
%\centerline{\includegraphics[width=0.8\columnwidth]{CAVIAR4REID-image}}
%\caption{The example images of two persons in CAVIAR4REID dataset.}
%\label{CAVIAR4REID_image}
%\end{center}
%\vskip -0in
%\vspace{-5mm}
%\end{figure}
\par
According to the evaluation protocol in \cite{li2013learning}, we randomly select 36 persons and use their images for training, and use the rest images for testing. For the testing images, we randomly select one image for each person to construct a probe set consisting of 36 images, and use the other test images as the gallery set. Finally, 10 partitions of training and test sets are constructed, and the final results are obtained by averaging over the 10 runs.
\par
\begin{figure}[ht]
\vskip 0in
\begin{center}
\centerline{\includegraphics[width=0.6\columnwidth]{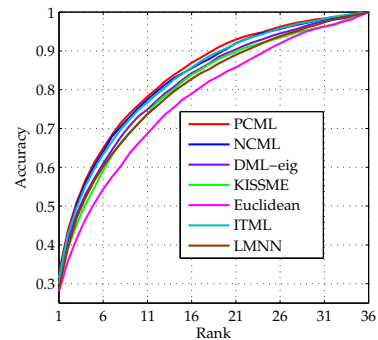}}
\caption{The CMC curves on the CAVIAR4REID dataset.}
\label{CAVIAR_CMC}
\end{center}
\vskip -0in
\vspace{-7mm}
\end{figure}

We report the CMC curves of PCML, NCML, DML-eig \cite{ying2012distance}, KISSME \cite{kostinger2012large}, ITML \cite{davis2007information}, LMNN \cite{weinberger2005LMNNNIPS} and Euclidean distance in Fig. \ref{CAVIAR_CMC}. One can see that PCML and NCML perform the best and the second best among all the competing methods, respectively. Table \ref{CAVIAR4REID_Rate} lists the re-identification accuracies and training time by different methods. PCML and NCML perform better than the other metric learning methods under all the ranks. We also report the training times of the competing metric learning methods in Table \ref{CAVIAR4REID_Rate}. PCML and NCML are much faster than the other metric learning methods except for KISSME.

\par

\begin{table}[!t]
\renewcommand{\arraystretch}{1.3}
\caption{Person re-identification accuracies (\%) and training time (s) on the CAVIAR4REID dataset.}
\label{CAVIAR4REID_Rate}
\vskip 0in
\begin{center}
\begin{tabular}{p{1.2cm}|m{0.9cm}m{0.9cm}m{1.1cm}m{1.1cm}|m{1.4cm}}
\hline
\multirow{2}{*}{\bfseries Methods} & \multicolumn{4}{c|}{\bfseries Accuracy (\%)} & \multirow{2}{1.4cm}{\bfseries Training Time (s)}\\
\cline{2-5}
& \bfseries Rank 1 & \bfseries Rank 5 & \bfseries Rank 10 & \bfseries Rank 15 \\
\hline
PCML & \textbf{32.86} & \textbf{61.26} & \textbf{76.06} & \textbf{85.34} & 11.47\\
NCML & 32.27 & 60.38 & 75.33 & 84.25 & 19.23\\
DML-eig\cite{ying2012distance} & 30.68 & 57.15 & 73.18 & 82.64 & 829.24\\
LMNN\cite{kostinger2012large} & 28.66  & 56.53  & 71.30  & 81.19 & 95.62\\
ITML\cite{kostinger2012large} & 31.48  & 59.56  & 74.83  & 84.15 & 2819.18\\
KISSME\cite{kostinger2012large} & 29.87  & 54.75  & 71.36  & 82.15 & 1.12\\
Euclidean & 27.98  & 50.67  & 66.25  & 77.54 & 0\\
\hline
\end{tabular}
\end{center}
\end{table}

\section{Conclusion}
\label{conclusion}
We proposed two distance metric learning models, namely Positive-semidefinite Constrained Metric Learning (PCML) and Nonnegative-coefficient Constrained Metric Learning (NCML). The proposed models can guarantee the positive semidefinite property of the learned matrix $\mathbf{M}$, and can be solved efficiently by the existing SVM solvers. Experimental results on nine UCI machine learning repository datasets and four handwritten digit datasets showed that, compared with the state-of-the-art metric learning methods, including NCA \cite{goldberger2004NCA}, ITML \cite{davis2007information}, MCML \cite{globerson2005MCML}, LDML \cite{guillaumin2009LDML}, LMNN \cite{weinberger2009LMNNJMLR}, PLML \cite{wang2012PLML}, and DML-eig \cite{ying2012distance}, the proposed PCML and NCML methods can not only achieve higher classification accuracy, but also are much faster in training. On average, they are 35 and 21 times faster than PLML, the 3rd fastest metric learning method, respectively. The experimental results on LFW, PubFig, VIPeR and CAVIAR4REID databases indicate that the proposed methods also perform very well in vision tasks such as face verification and person re-identification, leading to higher verification rates and very competitive training efficiency.

\appendices
\section{The Dual of PCML}
The original problem of PCML is formulated as
\begin{equation}
\begin{aligned}
  & \underset{\mathbf{M},b,\boldsymbol{\xi }}{\mathop{\min }}\,\quad \frac{1}{2}\left\| \mathbf{M} \right\|_{F}^{2}+C\sum\nolimits_{i,j}{{{\xi }_{ij}}} \\
 & \text{s}\text{.t}\text{.}\quad {{h}_{ij}}\left( \left\langle \mathbf{M},{{\mathbf{X}}_{ij}} \right\rangle +b \right)\ge 1-{{\xi }_{ij}}, {{\xi }_{ij}}\ge 0,\ \forall i,j \\
 & \quad \quad \mathbf{M}\succcurlyeq 0.
\end{aligned}
\end{equation}
Its Lagrangian is:
\begin{equation}
\begin{aligned}
  & L\left( \boldsymbol{\lambda },\boldsymbol{\kappa },\mathbf{Y},\mathbf{M},b,\boldsymbol{\xi } \right)=\frac{1}{2}\left\| \mathbf{M} \right\|_{F}^{2}+C\sum\nolimits_{i,j}{{{\xi }_{ij}}}\\
  & -\sum\nolimits_{i,j}{{{\lambda }_{ij}}\left[ {{h}_{ij}}\left( \left\langle \mathbf{M},{{\mathbf{X}}_{ij}} \right\rangle +b \right)-1+{{\xi }_{ij}} \right]} \\
 & -\sum\nolimits_{i,j}{{{\kappa }_{ij}}{{\xi }_{ij}}}-\left\langle \mathbf{Y},\mathbf{M} \right\rangle,
\end{aligned}
\end{equation}
where $\mathbf{\lambda }$, $\mathbf{\kappa }$ and $\mathbf{Y}$ are the Lagrange multipliers which satisfy ${{\lambda }_{ij}}\ge 0$, ${{\kappa }_{ij}}\ge 0, \forall i,j$, and $\mathbf{Y}\succcurlyeq 0$. Converting the original problem to its dual problem needs the following KKT conditions:
\begin{equation}
\begin{aligned}
\frac{\partial L\left( \boldsymbol{\lambda },\boldsymbol{\kappa },\mathbf{Y},\mathbf{M},b,\boldsymbol{\xi } \right)}{\partial \mathbf{M}}=\mathbf{0}
\Rightarrow \mathbf{M}-\sum\nolimits_{i,j}{{{\lambda }_{ij}}{{h}_{ij}}{{\mathbf{X}}_{ij}}}-\mathbf{Y}=\mathbf{0},
\end{aligned}
\end{equation}
\begin{equation}
\begin{aligned}
\frac{\partial L\left( \boldsymbol{\lambda },\boldsymbol{\kappa },\mathbf{Y},\mathbf{M},b,\boldsymbol{\xi } \right)}{\partial b}=0\Rightarrow \sum\nolimits_{i,j}{{{\lambda }_{ij}}{{h}_{ij}}}=0,
\end{aligned}
\end{equation}
\begin{equation}
\begin{aligned}
\frac{\partial L\left( \boldsymbol{\lambda },\boldsymbol{\kappa },\mathbf{Y},\mathbf{M},b,\boldsymbol{\xi } \right)}{\partial {{\xi }_{ij}}}=C-{{\lambda }_{ij}}-{{\kappa }_{ij}}=0\Rightarrow \\
0\le {{\lambda }_{ij}}\le C,\ \forall i,j,
\end{aligned}
\end{equation}
\begin{equation}
\begin{aligned}
{{h}_{ij}}\left( \left\langle \mathbf{M},{{\mathbf{X}}_{ij}} \right\rangle +b \right)-1+{{\xi }_{ij}}\ge 0,\ {{\xi }_{ij}}\ge 0,
\end{aligned}
\end{equation}
\begin{equation}
\begin{aligned}
{{\lambda }_{ij}}\ge 0,\ {{\kappa }_{ij}}\ge 0,\ \mathbf{Y}\succcurlyeq 0,
\end{aligned}
\end{equation}
\begin{equation}
\begin{aligned}
{{\lambda }_{ij}}\left[ {{h}_{ij}}\left( \left\langle \mathbf{M},{{\mathbf{X}}_{ij}} \right\rangle +b \right)-1+{{\xi }_{ij}} \right]=0,\ {{\kappa }_{ij}}{{\xi }_{ij}}=0.
\end{aligned}
\end{equation}
Equation (37) implies the following relationship between $\boldsymbol{\lambda }$, $\mathbf{Y}$ and $\mathbf{M}$:
\begin{equation}
\begin{aligned}
\mathbf{M}=\sum\nolimits_{i,j}{{{\lambda }_{ij}}{{h}_{ij}}{{\mathbf{X}}_{ij}}}+\mathbf{Y}.
\end{aligned}
\end{equation}
Substituting (37)$\sim$(39) back into the Lagrangian, we get the following Lagrange dual problem of PCML:
\begin{equation}
\begin{aligned}
  & \underset{\mathbf{\lambda },\mathbf{Y}}{\mathop{\max }}\,\quad -\frac{1}{2}\left\| \sum\nolimits_{i,j}{{{\lambda }_{ij}}{{h}_{ij}}{{\mathbf{X}}_{ij}}}+\mathbf{Y} \right\|_{F}^{2}+\sum\nolimits_{i,j}{{{\lambda }_{ij}}} \\
 & \text{         s}\text{.t}\text{.}\quad \sum\nolimits_{i,j}{{{\lambda }_{ij}}{{h}_{ij}}}=0, 0\le {{\lambda }_{ij}}\le C, \forall i,j,\quad \mathbf{Y}\succcurlyeq 0.
\end{aligned}
\end{equation}
\par
As we can see from (43) and (44), $\mathbf{M}$ is explicitly determined by the training procedure, but $b$ is not. Nevertheless, $b$ can be easily found by using the KKT complementarity condition in (39) and (42), which show that ${{\xi }_{ij}}=0$ if ${{\lambda }_{ij}}<C$, and ${{h}_{ij}}\left( \left\langle \mathbf{M},{{\mathbf{X}}_{ij}} \right\rangle +b \right)-1+{{\xi }_{ij}}=0$ if ${{\lambda }_{ij}}>0$. Thus we can simply take any training point, for which $0<{{\lambda }_{ij}}<C$, to compute $b$ by
\begin{equation}
\begin{aligned}
b=\frac{1}{{{h}_{ij}}}-\left\langle \mathbf{M},{{\mathbf{X}}_{ij}} \right\rangle ,\quad \text{for}\ \text{all}\ 0<{{\lambda }_{ij}}<C.
\end{aligned}
\end{equation}
Note that it is numerically wiser to take the average over all such training data points to compute $b$. After $b$ is computed, we can compute ${{\xi }_{ij}}$ by
\begin{equation}
{{\xi }_{ij}}=\left\{ \begin{aligned}
  & 0,\quad \text{for all}\ {{\lambda }_{ij}}<C \\
  &{{\left[ 1-{{h}_{ij}}\left( \left\langle \mathbf{M},{{\mathbf{X}}_{ij}} \right\rangle +b \right) \right]}_{+}},\quad \text{for all}\ {{\lambda }_{ij}}=C,
 \end{aligned} \right.
\end{equation}
where the term ${{\left[ z \right]}_{+}}=\max \left( z,0 \right)$ denotes the standard hinge loss.
% you can choose not to have a title for an appendix
% if you want by leaving the argument blank
\section{The Dual of NCML}
The original problem of NCML is as follows:
\begin{equation}
\begin{aligned}
  & \underset{\boldsymbol{\alpha },b,\boldsymbol{\xi }}{\mathop{\min }}\,\quad \frac{1}{2}\sum\nolimits_{i,j}{\sum\nolimits_{k,l}{{{\alpha }_{ij}}{{\alpha }_{kl}}\left\langle {{\mathbf{X}}_{ij}},{{\mathbf{X}}_{kl}} \right\rangle }}+C\sum\nolimits_{i,j}{{{\xi }_{ij}}} \\
 & \text{s}\text{.t}\text{.}\quad {{h}_{ij}}\left( \sum\nolimits_{k,l}{{{\alpha }_{kl}}\left\langle {{\mathbf{X}}_{ij}},{{\mathbf{X}}_{kl}} \right\rangle }+b \right)\ge 1-{{\xi }_{ij}} \\
 & \quad \quad {{\xi }_{ij}}\ge 0, {{\alpha }_{ij}}\ge 0,\ \forall i,j.
\end{aligned}
\end{equation}
Its Lagrangian can be defined as:
\begin{equation}
\begin{aligned}
  & L\left( \boldsymbol{\beta },\boldsymbol{\sigma },\boldsymbol{\nu },\boldsymbol{\alpha },b,\boldsymbol{\xi } \right)=\frac{1}{2}\sum\nolimits_{i,j,k,l}{{{{\alpha }_{ij}}{{\alpha }_{kl}}\left\langle {{\mathbf{X}}_{ij}},{{\mathbf{X}}_{kl}} \right\rangle }}+C\sum\nolimits_{i,j}{{{\xi }_{ij}}} \\
 & -\sum\nolimits_{i,j}{{{\beta }_{ij}}\left[ {{h}_{ij}}\left( \sum\nolimits_{kl}{{{\alpha }_{kl}}\left\langle {{\mathbf{X}}_{ij}},{{\mathbf{X}}_{kl}} \right\rangle }+b \right)-1+{{\xi }_{ij}} \right]}\\
 & -\sum\nolimits_{i,j}{{{\nu }_{ij}}{{\xi }_{ij}}}-\sum\nolimits_{i,j}{{{\sigma }_{ij}}{{\alpha }_{ij}}},
\end{aligned}
\end{equation}
where $\mathbf{\beta }$, $\mathbf{\sigma }$ and $\mathbf{\nu }$ are the Lagrange multipliers which satisfy ${{\beta }_{ij}}\ge 0$, ${{\sigma }_{ij}}\ge 0$ and ${{\nu }_{ij}}\ge 0$, $\forall i,j$. Converting the original problem to its dual problem needs the following KKT conditions:
\begin{equation}
\begin{aligned}
& \frac{\partial L\left( \boldsymbol{\beta },\boldsymbol{\sigma },\boldsymbol{\nu },\boldsymbol{\alpha },b,\boldsymbol{\xi } \right)}{\partial {{\alpha }_{ij}}}=0\Rightarrow \\
& \sum\nolimits_{k,l}{{{\alpha }_{kl}}\left\langle {{\mathbf{X}}_{ij}},{{\mathbf{X}}_{kl}} \right\rangle }-\sum\nolimits_{k,l}{{{\beta }_{kl}}{{h}_{kl}}\left\langle {{\mathbf{X}}_{ij}},{{\mathbf{X}}_{kl}} \right\rangle }-{{\sigma }_{ij}}=0,
\end{aligned}
\end{equation}
\begin{equation}
\begin{aligned}
\frac{\partial L\left( \boldsymbol{\beta },\boldsymbol{\sigma },\boldsymbol{\nu },\boldsymbol{\alpha },b,\boldsymbol{\xi } \right)}{\partial b}=0\Rightarrow \sum\nolimits_{i,j}{{{\beta }_{ij}}{{h}_{ij}}}=0,
\end{aligned}
\end{equation}
\begin{equation}
\begin{aligned}
& \frac{\partial L\left( \boldsymbol{\beta },\boldsymbol{\sigma },\boldsymbol{\nu },\boldsymbol{\alpha },b,\boldsymbol{\xi } \right)}{\partial {{\xi }_{ij}}}=0\Rightarrow C-{{\beta }_{ij}}-{{\nu }_{ij}}=0 \Rightarrow\\
& 0\le {{\beta }_{ij}}\le C,
\end{aligned}
\end{equation}
\begin{equation}
\begin{aligned}
& {{h}_{ij}}\left( \sum\nolimits_{k,l}{{{\alpha }_{kl}}\left\langle {{\mathbf{X}}_{ij}},{{\mathbf{X}}_{kl}} \right\rangle }+b \right)-1+{{\xi }_{ij}}\ge 0,\\
&\quad\quad\quad\quad\quad {{\xi }_{ij}}\ge 0,\ {{\alpha }_{ij}}\ge 0,\ \forall i,j,
\end{aligned}
\end{equation}
\begin{equation}
\begin{aligned}
{{\beta }_{ij}}\ge 0,\ {{\sigma }_{ij}}\ge 0,\ {{\nu }_{ij}}\ge 0,\quad \forall i,j,
\end{aligned}
\end{equation}
\begin{equation}
\begin{aligned}
& {{\beta }_{ij}}\left[ {{h}_{ij}}\left( \sum\nolimits_{k,l}{{{\alpha }_{kl}}\left\langle {{\mathbf{X}}_{ij}},{{\mathbf{X}}_{kl}} \right\rangle }+b \right)-1+{{\xi }_{ij}} \right]=0, \\
&\quad\quad\quad\quad\quad {{\nu }_{ij}}{{\xi }_{ij}}=0,\ {{\sigma }_{ij}}{{\alpha }_{ij}}=0,\quad \forall i,j.
\end{aligned}
\end{equation}
Here we introduce a coefficient vector $\boldsymbol{\eta }$, which satisfies ${{\sigma }_{ij}}=\sum\nolimits_{k,l}{{{\eta }_{kl}}\left\langle {{\mathbf{X}}_{ij}},{{\mathbf{X}}_{kl}} \right\rangle }$. Note that $\left\langle {{\mathbf{X}}_{ij}},{{\mathbf{X}}_{kl}} \right\rangle $ is a positive definite kernel. So we can guarantee that every $\boldsymbol{\eta }$ corresponds to a unique $\boldsymbol{\sigma }$, and vice versa. Equation (49) implies the following relationship between $\boldsymbol{\alpha }$, $\boldsymbol{\beta }$ and $\boldsymbol{\eta }$:
\begin{equation}
\begin{aligned}
{{\alpha }_{ij}}={{\beta }_{ij}}{{h}_{ij}}+{{\eta }_{ij}},\quad \forall i,j.
\end{aligned}
\end{equation}
Substituting (49)$\sim$(51) back into the Lagrangian, we get the Lagrange dual problem of NCML as follows:
\begin{equation}
\begin{aligned}
  & \underset{\boldsymbol{\eta },\boldsymbol{\beta }}{\mathop{\max }}\, -\frac{1}{2}\sum\limits_{i,j,k,l}{{\left( {{\beta }_{ij}}{{h}_{ij}}+{{\eta }_{ij}} \right)\left( {{\beta }_{kl}}{{h}_{kl}}+{{\eta }_{kl}} \right)\left\langle {{\mathbf{X}}_{ij}},{{\mathbf{X}}_{kl}} \right\rangle }} \\
  & +\sum\limits_{i,j}{{{\beta }_{ij}}} \\
 & \text{s}\text{.t}\text{.}\quad \sum\nolimits_{k,l}{{{\eta }_{kl}}\left\langle {{\mathbf{X}}_{ij}},{{\mathbf{X}}_{kl}} \right\rangle }\ge 0, 0\le {{\beta }_{ij}}\le C,\ \forall i,j \\
 & \quad \quad \sum\nolimits_{i,j}{{{\beta }_{ij}}{{h}_{ij}}}=0.
\end{aligned}
\end{equation}
\par
Analogous to PCML, we can use the KKT complementarity condition in (50) to compute $b$ and ${{\xi }_{ij}}$ in NCML. Equations (51) and (54) show that ${{\xi }_{ij}}=0$ if ${{\beta }_{ij}}<C$, and ${{h}_{ij}}\left( \sum\nolimits_{kl}{{{\alpha }_{kl}}\left\langle {{\mathbf{X}}_{ij}},{{\mathbf{X}}_{kl}} \right\rangle }+b \right)-1+{{\xi }_{ij}}=0$ if ${{\beta }_{ij}}>0$. Thus we can simply take any training data point, for which $0<{{\beta }_{ij}}<C$, to compute $b$ by
\begin{equation}
\begin{aligned}
b=\frac{1}{{{h}_{ij}}}-\sum\nolimits_{k,l}{{{\alpha }_{kl}}\left\langle {{\mathbf{X}}_{ij}},{{\mathbf{X}}_{kl}} \right\rangle }.
\end{aligned}
\end{equation}
After obtain $b$, we can compute ${{\beta }_{ij}}$ by
\begin{equation}
{{\xi }_{ij}}=\left\{ \begin{aligned}
  & 0,\forall\ {{\beta }_{ij}}<C \\
 & {{\left[ 1-{{h}_{ij}}\left( \sum\nolimits_{k,l}{{{\alpha }_{kl}}\left\langle {{\mathbf{X}}_{ij}},{{\mathbf{X}}_{kl}} \right\rangle }+b \right) \right]}_{+}},\forall\ {{\beta }_{ij}}=C,
\end{aligned}\right.
\end{equation}
where the term ${{\left[ z \right]}_{+}}=\max \left( z,0 \right)$ denotes the standard hinge loss.
\section{The Dual of the Subproblem on $\boldsymbol{\eta}$ in NCML}
The subproblem on $\boldsymbol{\eta }$ is formulated as follows:
\begin{equation}
\begin{aligned}
  & \underset{\boldsymbol{\eta }}{\mathop{\min }}\,\quad \frac{1}{2}\sum\nolimits_{i,j}{\sum\nolimits_{k,l}{{{\eta }_{ij}}{{\eta }_{kl}}\left\langle {{\mathbf{X}}_{ij}},{{\mathbf{X}}_{kl}} \right\rangle }}+\sum\nolimits_{i,j}{{{\eta }_{ij}}{{\gamma }_{ij}}} \\
 & \text{s}\text{.t}\text{.}\quad \sum\nolimits_{k,l}{{{\eta }_{kl}}\left\langle {{\mathbf{X}}_{ij}},{{\mathbf{X}}_{kl}} \right\rangle }\ge 0,\ \forall i,j,
\end{aligned}
\end{equation}
where ${{\gamma }_{ij}}=\sum\nolimits_{k,l}{{{\beta }_{kl}}{{h}_{kl}}\left\langle {{\mathbf{X}}_{ij}},{{\mathbf{X}}_{kl}} \right\rangle }$. Its Lagrangian is:
\begin{equation}
\begin{aligned}
& L\left( \boldsymbol{\mu },\boldsymbol{\eta } \right)=\frac{1}{2}\sum\nolimits_{i,j}{\sum\nolimits_{k,l}{{{\eta }_{ij}}{{\eta }_{kl}}\left\langle {{\mathbf{X}}_{ij}},{{\mathbf{X}}_{kl}} \right\rangle }}\\
& +\sum\nolimits_{i,j}{{{\eta }_{ij}}{{\gamma }_{ij}}}-\sum\nolimits_{i,j}{{{\mu }_{ij}}\sum\nolimits_{k,l}{{{\eta }_{kl}}\left\langle {{\mathbf{X}}_{ij}},{{\mathbf{X}}_{kl}} \right\rangle }},
\end{aligned}
\end{equation}
where $\boldsymbol{\mu }$ is the Lagrange multiplier which satisfies ${{\mu }_{ij}}\ge 0, \forall i,j$. Converting the original problem to its dual problem needs the following KKT condition:
\begin{equation}
\begin{aligned}
& \frac{\partial L\left( \boldsymbol{\mu },\boldsymbol{\eta } \right)}{\partial {{\eta }_{ij}}}=0\Rightarrow \\
& \sum\nolimits_{k,l}{{{\eta }_{kl}}\left\langle {{\mathbf{X}}_{ij}},{{\mathbf{X}}_{kl}} \right\rangle }+{{\gamma }_{ij}}-\sum\nolimits_{k,l}{{{\mu }_{kl}}\left\langle {{\mathbf{X}}_{ij}},{{\mathbf{X}}_{kl}} \right\rangle }=0.
\end{aligned}
\end{equation}
Equation (61) implies the following relationship between $\boldsymbol{\mu }$, $\boldsymbol{\eta }$ and $\boldsymbol{\beta }$:
\begin{equation}
\begin{aligned}
{{\eta }_{ij}}={{\mu }_{ij}}-{{h}_{ij}}{{\beta }_{ij}},\quad \forall i,j.
\end{aligned}
\end{equation}
Substituting (61) and (62) back into the Lagrangian, we get the following Lagrange dual problem of the subproblem on $\boldsymbol{\eta }$:
\begin{equation}
\begin{aligned}
  & \underset{\boldsymbol{\mu }}{\mathop{\max }}\,\quad -\frac{1}{2}\sum\nolimits_{i,j}{\sum\nolimits_{k,l}{{{\mu }_{ij}}{{\mu }_{kl}}\left\langle {{\mathbf{X}}_{ij}},{{\mathbf{X}}_{kl}} \right\rangle }}+\sum\nolimits_{i,j}{{{\gamma }_{ij}}{{\mu }_{ij}}}\\
  & -\frac{1}{2}\sum\nolimits_{i,j}{\sum\nolimits_{k,l}{{{\beta }_{ij}}{{\beta }_{kl}}{{h}_{ij}}{{h}_{kl}}\left\langle {{\mathbf{X}}_{ij}},{{\mathbf{X}}_{kl}} \right\rangle }} \\
 & \text{s}\text{.t}\text{.}\quad {{\mu }_{ij}}\ge 0,\forall i,j.
\end{aligned}
\end{equation}
Since $\boldsymbol{\beta }$ is fixed in this subproblem, $\sum\nolimits_{i,j}{\sum\nolimits_{k,l}{{{\beta }_{ij}}{{\beta }_{kl}}{{h}_{ij}}{{h}_{kl}}\left\langle {{\mathbf{X}}_{ij}},{{\mathbf{X}}_{kl}} \right\rangle }}$ remains constant in (63). Thus we can omit this term and have the following simplified Lagrange dual problem:
\begin{equation}
\begin{aligned}
  & \underset{\boldsymbol{\mu }}{\mathop{\max }}\,\quad -\frac{1}{2}\sum\nolimits_{i,j}{\sum\nolimits_{k,l}{{{\mu }_{ij}}{{\mu }_{kl}}\left\langle {{\mathbf{X}}_{ij}},{{\mathbf{X}}_{kl}} \right\rangle }}+\sum\nolimits_{i,j}{{{\gamma }_{ij}}{{\mu }_{ij}}} \\
 & \text{s}\text{.t}\text{.}\quad {{\mu }_{ij}}\ge 0,\forall i,j.
\end{aligned}
\end{equation}

% use section* for acknowledgement

% use section* for acknowledgement
%\ifCLASSOPTIONcompsoc
  % The Computer Society usually uses the plural form
%  \section*{Acknowledgments}
%  This work was supported in part by the National Natural Science Foundation of China under Grant 61271093 and Grant 61001037, and the Program of Ministry of Education for New Century Excellent Talents under Grant NCET-12-0150.
%\else
  % regular IEEE prefers the singular form
%  \section*{Acknowledgment}
%\fi

% Can use something like this to put references on a page
% by themselves when using endfloat and the captionsoff option.
\ifCLASSOPTIONcaptionsoff
  \newpage
\fi

% trigger a \newpage just before the given reference
% number - used to balance the columns on the last page
% adjust value as needed - may need to be readjusted if
% the document is modified later
%\IEEEtriggeratref{8}
% The "triggered" command can be changed if desired:
%\IEEEtriggercmd{\enlargethispage{-5in}}

% references section

% can use a bibliography generated by BibTeX as a .bbl file
% BibTeX documentation can be easily obtained at:
% http://www.ctan.org/tex-archive/biblio/bibtex/contrib/doc/
% The IEEEtran BibTeX style support page is at:
% http://www.michaelshell.org/tex/ieeetran/bibtex/
%\bibliographystyle{IEEEtran}
% argument is your BibTeX string definitions and bibliography database(s)
%\bibliography{IEEEabrv,../bib/paper}
%
% <OR> manually copy in the resultant .bbl file
% set second argument of \begin to the number of references
% (used to reserve space for the reference number labels box)
%\begin{thebibliography}{1}

%\bibitem{IEEEhowto:kopka}
%H.~Kopka and P.~W. Daly, \emph{A Guide to \LaTeX}, 3rd~ed.\hskip 1em plus
%  0.5em minus 0.4em\relax Harlow, England: Addison-Wesley, 1999.

%\end{thebibliography}
\bibliographystyle{IEEEtran}
% argument is your BibTeX string definitions and bibliography database(s)
%\bibliography{IEEEabrv,../bib/paper}
\bibliography{bare_jrnl_compsoc}
% biography section
%
% If you have an EPS/PDF photo (graphicx package needed) extra braces are
% needed around the contents of the optional argument to biography to prevent
% the LaTeX parser from getting confused when it sees the complicated
% \includegraphics command within an optional argument. (You could create
% your own custom macro containing the \includegraphics command to make things
% simpler here.)
%\begin{IEEEbiography}[{\includegraphics[width=1in,height=1.25in,clip,keepaspectratio]{mshell}}]{Michael Shell}
% or if you just want to reserve a space for a photo:

%\begin{IEEEbiography}{Michael Shell}
%Biography text here.
%\end{IEEEbiography}

% if you will not have a photo at all:
%\begin{IEEEbiographynophoto}{John Doe}
%Biography text here.
%\end{IEEEbiographynophoto}

% insert where needed to balance the two columns on the last page with
% biographies
%\newpage

%\begin{IEEEbiographynophoto}{Jane Doe}
%Biography text here.
%\end{IEEEbiographynophoto}

% You can push biographies down or up by placing
% a \vfill before or after them. The appropriate
% use of \vfill depends on what kind of text is
% on the last page and whether or not the columns
% are being equalized.

%\vfill

% Can be used to pull up biographies so that the bottom of the last one
% is flush with the other column.
%\enlargethispage{-5in}

% that's all folks
\end{document}